%% file: main.tex
\documentclass[11pt]{article}

\usepackage[preprint]{acl}

\usepackage{times}
\usepackage{latexsym}
\usepackage[T1]{fontenc}
\usepackage[utf8]{inputenc}
\usepackage{microtype}
\usepackage{inconsolata}

\usepackage{graphicx}
\usepackage{amsmath}
\usepackage{booktabs}
\usepackage{adjustbox}
\usepackage{caption}
\usepackage[most]{tcolorbox}
\usepackage[nameinlink]{cleveref}
\usepackage{float}
\usepackage{gensymb}
\usepackage{stfloats}
\usepackage{tabularx}
\usepackage{enumitem}
\usepackage[hang,flushmargin]{footmisc}
\usepackage[framemethod=TikZ]{mdframed}
\usepackage{xspace}

\usepackage{xcolor}

\renewcommand{\paragraph}[1]{\vspace{2pt}\noindent\textbf{#1}\;}

\definecolor{lightYellow}{RGB}{254,254,233}
\definecolor{midGray}{gray}{0.40}

\mdfdefinestyle{takeawayFrame}{%
  linecolor=midGray, linewidth=0.5pt, roundcorner=3pt,
  skipabove=7pt, skipbelow=0pt,
  innertopmargin=5pt, innerbottommargin=5pt,
  innerrightmargin=7pt, innerleftmargin=7pt,
  leftmargin=0pt, rightmargin=0pt,
  backgroundcolor=lightYellow,
  userdefinedwidth=\linewidth}

\mdfdefinestyle{citationFrame}{%
  linecolor=midGray, linewidth=0.5pt, roundcorner=3pt,
  skipabove=5pt, skipbelow=0pt,
  innertopmargin=5pt, innerbottommargin=5pt,
  innerrightmargin=7pt, innerleftmargin=7pt,
  leftmargin=0pt, rightmargin=0pt,
  backgroundcolor=lightYellow}

\newcommand{\fp}{\textsc{Facet-Probe}\xspace}
\newcommand{\odi}{\textsc{ODI}\xspace}
\newcommand{\mmlupro}{{\sc{mmlu-pro}}\xspace}
\newcommand{\medxpertqa}{{\sc{medxpertqa}}\xspace}
\newcommand{\mathvision}{{\sc{mathvision}}\xspace}
\newcommand{\mantiseval}{{\sc{mantis-eval}}\xspace}
\newcommand{\medframeqa}{{\sc{medframeqa}}\xspace}
\newcommand{\hotpotqa}{{\sc{hotpotqa}}\xspace}
\newcommand{\musique}{{\sc{musique}}\xspace}
\newcommand{\multihoprag}{{\sc{multihop-rag}}\xspace}
\newcommand{\gsm}{{\sc{gsm8k}}\xspace}
\newcommand{\humaneval}{{\sc{humaneval}}\xspace}
\newcommand{\csqa}{{\sc{commonsenseqa}}\xspace}

\newcommand{\multichallenge}{{\sc{multichallenge}}\xspace}

\newcommand{\optionorder}{{\sc{option-order}}\xspace}
\newcommand{\evchunkorder}{{\sc{evidence-chunk-order}}\xspace}
\newcommand{\docrankorder}{{\sc{document-rank-order}}\xspace}
\newcommand{\imagesetorder}{{\sc{image-set-order}}\xspace}
\newcommand{\mmorder}{{\sc{mixed-modality-order}}\xspace}

\IfFileExists{draft_figures_latest/writing_extras/tex_snippets.tex}{%
  \input{draft_figures_latest/writing_extras/tex_snippets}%
}{}

\providecommand{\fpRatioSigmaImageOpt}{1.73}
\providecommand{\fpRatioDeltaImageOpt}{8.25}
\providecommand{\fpNModels}{18}
\providecommand{\fpNDatasets}{12}  
\providecommand{\fpNTrials}{${>}400{,}000$}
\providecommand{\fpBestFlip}{13.4\%}    
\providecommand{\fpFrontierSubsetMean}{17\%}   
\providecommand{\fpOpenWeightSubsetMean}{36\%} 

\title{Same Evidence, Different Answer: \\Auditing Order Sensitivity in Multimodal Large Language Models}

\author{%
 Akshay Paruchuri \quad Sanmi Koyejo \quad Ehsan Adeli \\
 Stanford University
}

\begin{document}
\maketitle
\begin{abstract}
Standard benchmarks for multimodal large language models (MLLMs) score each item on one canonical ordering and miss whether order-irrelevant shuffling changes the answer, a baseline reliability property called for by emerging AI evaluation guidelines. We introduce \textsc{Facet-Probe}, a five-facet audit (option, evidence-chunk, document-rank, image-set, and mixed-modality ordering) of 18 frontier and open-weight MLLMs. A Bayesian item-response model separates ordering noise from per-facet bias; a same-ordering control estimates the decoder-stochastic floor for observed flips.

\vspace{1pt}
\noindent\textbf{Findings.}
None of the 18 MLLMs we audit are order-invariant: screened per-facet panel-mean flip rates span 24--50\%. A Gemini same-ordering control at temperature~0 estimates a substantial ordering excess over a same-input decoder-noise floor in verified cells. Capability predicts but does not eliminate flips; the best model still flips on \fpBestFlip{} of trials.

\vspace{1pt}
\noindent\textbf{Implications.}
In our Gemini mitigation tests, training-free prompt changes are modality-conditional and do not transfer from text to visual reasoning. These results suggest that prompt-level mitigation alone is unlikely to provide general order robustness, motivating future work on training-time and architectural approaches. We propose cross-ordering flip rate as a standard reporting axis for MLLMs.

\vspace{1pt}
\noindent\textbf{Code and Audit Artifacts:} \url{https://github.com/yahskapar/facet-probe}
\end{abstract}

\input{sections/1-introduction}
\input{sections/2-related_work}
\input{sections/3-methods}
\input{sections/4-findings}
\input{sections/5-discussion}
\clearpage
\input{sections/6-limitations}

\clearpage


\bibliography{references}

\clearpage
\appendix

\input{supplementary/sections/1-extended_dataset_details}
\input{supplementary/sections/2-irt_methodology}
\input{supplementary/sections/3-per_facet_per_model_tables}
\input{supplementary/sections/4-robustness_analysis}
\input{supplementary/sections/5-llm_judge_methodology}
\input{supplementary/sections/6-additional_facet_results}
\input{supplementary/sections/7-mitigation_extended}
\input{supplementary/sections/8-mixed_modality_facet}
\input{supplementary/sections/9-mechanism_extended}
\input{supplementary/sections/10-extensibility}

\end{document}

%% file: sections/1-introduction.tex
\section{Introduction}
\label{sec:intro}

\begin{figure}[!t]
    \centering
    \includegraphics[width=\columnwidth]{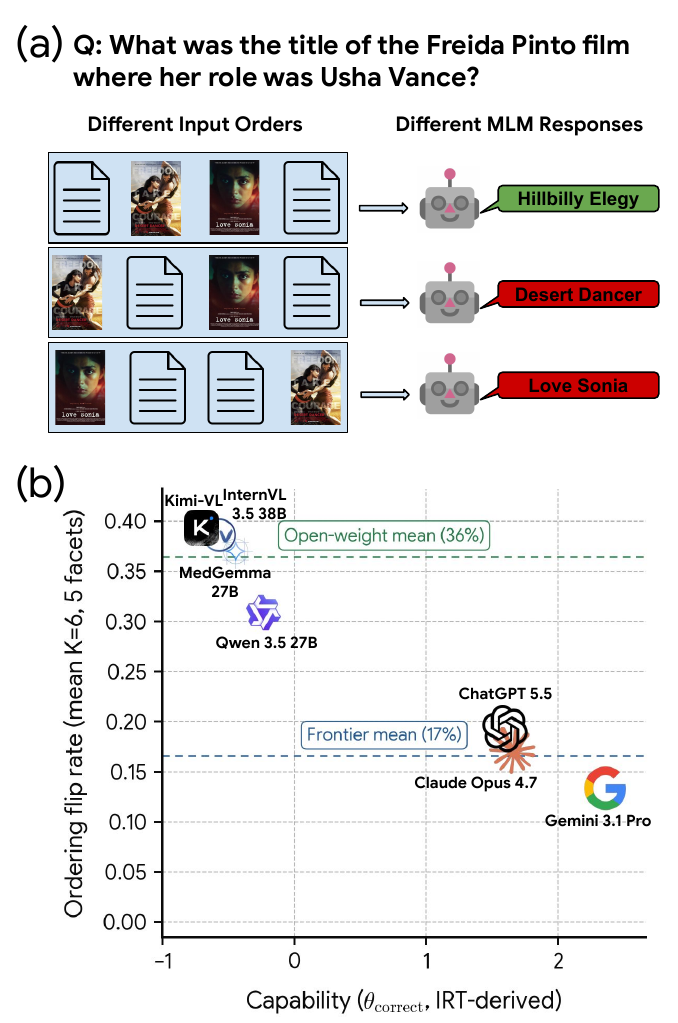}
    \caption{\textbf{MLLMs Remain Order-Sensitive.}
    \textbf{(a)}~The same evidence, presented in three
    different, yet semantically equivalent, orderings,
    yields three different answers; only one is correct.
	    \textbf{(b)}~Capability vs.\ mean $K{=}6$ flip rate over the 5
	    facets, shown for 7 best-of-family models. Capability predicts
	    but does not eliminate flips ($\rho \approx -0.95$ on the full
	    \fpNModels-model panel); the best model still flips on
	    \fpBestFlip{} of trials.}
    \label{fig:teaser}
\end{figure}

Standard benchmarks for multimodal large language models (MLLMs)
score each item on a single canonical ordering and implicitly treat
semantically equivalent permutations as interchangeable
\citep{wang2024mmlu, zuo2025medxpertqa, jiang2024mantis}. In deployed
systems, input order is set by retrieval pipelines
\citep{lewis2020retrieval, abootorabi2025ask, yao2025spotlight},
agentic tool schedulers \citep{shinn2023reflexion, faghih2025tool,
mei2025survey}, or clinical and consumer-facing interfaces
\citep{haberle2024impact, tierney2024ambient, hurt2025use,
paruchuri2025s}, not by the end user. If a model's answer changes
under those permutations, canonical benchmark accuracy overstates
reliability for multimodal RAG, agentic tool use, and clinical or
other high-stakes settings where users reasonably expect the same
evidence to yield the same answer.

\paragraph{MLLMs are not order-invariant.}
The same query under three semantically equivalent input orderings
yields three different answers (\Cref{fig:teaser}). The behavior is
not confined to small or older models and is not a measurement
artifact of any single benchmark. Emerging AI evaluation guidance
asks that benchmarks ``qualify'' the order-irrelevance assumption
explicitly under input perturbations \citep{salaudeen2025measurement,
bean2026measuring, nist800-2}, but no evaluation in widespread use
does so for multimodal ordering.

\paragraph{Prior work touches this but does not close it.}
\citet{tan2024order} measures multimodal positional bias on a single
ordering facet across 5--8 pre-2025 multimodal models. Promptception
\citep{ismithdeen2025promptception} characterizes prompt-template sensitivity on
10 multimodal models but does not study orthogonal ordering facets or the
2026 frontier generation. M3IRT \citep{uebayashi2026evaluating} applies
item-response theory to multimodal evaluation along the cross-modal
shortcut axis, not the ordering-noise axis. \citet{kostiuk2026one}
establish ranking instability on text-only and embedding-only evaluation
but do not study multimodal ordering or estimate the ordering excess over
same-input decoder-stochastic flips. Foundational work on option-order and
dialog-turn sensitivity in text-only LMs
\citep{pezeshkpour2024large, laban2023you, chen2024premise} sets up
the problem we extend to the multimodal frontier. No prior work
runs a same-ordering control that estimates the ordering excess over a
same-input decoder-noise floor; without
that decomposition, published flip-rate numbers are ambiguous.
Therefore, this paper asks the following:

\begin{mdframed}[style=citationFrame]
\bfseries
Do multimodal large language models give the same answer when the same evidence
is presented in a different order? If not, how prevalent is the failure,
and which interventions reduce it?
\end{mdframed}\vspace{-3pt}

\paragraph{This paper introduces \fp.}
\fp is a five-facet audit of \fpNModels{} multimodal large language models
(6 frontier closed-source: Gemini-Pro 3.1, Gemini-Flash 3, Claude Opus 4.7,
Claude Sonnet 4.6, ChatGPT 5.5, ChatGPT 5.4-mini; 12 open-weight: Qwen3.5,
InternVL3.5, Kimi-VL, and MedGemma families) over \fpNDatasets{} datasets and
\fpNTrials{} trials, paired with \odi (Ordering-Decomposed Item-Response
Theory), a 2PL Bayesian item-response decomposition that separates
per-item ordering noise $\sigma_\pi$ from per-facet systematic bias
$|\delta|$. Capability predicts but does not
eliminate flips (Spearman $\rho \approx -0.95$); the best model flips on
\fpBestFlip{} of trials. In our Gemini mitigation tests, training-free
prompt changes are modality-conditional and do not transfer from text
to visual reasoning.
\textbf{Our contributions are as follows:}

\begin{itemize}[topsep=0pt, leftmargin=15pt, itemsep=2pt]
  \item \Cref{sec:cross_facet_panel}: \fp, a five-facet ordering audit
        of \fpNModels{} multimodal LLMs across \fpNDatasets{} datasets, decomposed
        with \odi (2PL Bayesian IRT). A same-ordering control estimates a
        substantial ordering excess over a same-input decoder-noise floor in verified Gemini cells at
        temperature~0; screened image-set claims are anchored on
        \medframeqa{} rather than position-referential \mantiseval{} items.
  \item \Cref{sec:calibration_and_decoupling}: Capability predicts but
        does not eliminate flips (panel-wide $\rho \approx -0.95$; best model
        still flips on \fpBestFlip{}); stated confidence under-predicts
        ordering risk by 9--62 pp; mechanism analysis recovers content
        rationalization as a common failure mode, with the original
        image-set mechanism row treated as exploratory because it was
        sampled before the \mantiseval{} position-reference screen.
  \item \Cref{sec:mitigation}: Training-free, prompt-level mitigations
        are modality-conditional and do not compose; we report a
        cost-Pareto frontier.
\end{itemize}

\noindent We plan to release the full evaluation code and related artifacts for further use by the research community.

%% file: sections/2-related_work.tex
\section{Related Work}
\label{sec:related_work}

\fp differs from prior work along five axes: \emph{input facets}
(option, evidence-chunk, document rank, image-set, mixed-modality),
\emph{model generation} (frontier and open-weight 2026 MLLMs),
\emph{measurement target} (cross-order instability), \emph{decomposition}
via a same-ordering control, and \emph{mitigation evaluation}. The
decomposition axis is central: no prior work estimates the ordering
excess over same-input decoder-stochastic flips on the 2026 multimodal
frontier.

\paragraph{Ordering sensitivity in language models.}
LMs are sensitive to many semantically irrelevant input perturbations:
few-shot demonstration order \citep{lu2022fantastically}, prompt format
\citep{sclar2024quantifying, mizrahi2024state}, multiple-choice option
order \citep{pezeshkpour2024large, zheng2024large}, position in long
context \citep{liu2024lost}, two-turn challenges \citep{laban2023you},
reasoning premise order \citep{chen2024premise}, embedding-prompt
instability \citep{kostiuk2026one}, and a theoretical impossibility for
inverse-permutation learning in decoder-only transformers
\citep{alur2025impossibility}. Multimodal work documents single-facet
positional bias \citep{tan2024order}, multi-image VLM position bias
\citep{tian2025identifying}, prompt-template sensitivity on 10 models
\citep{ismithdeen2025promptception}, and U-shaped position bias in
multimodal RAG \citep{yao2025spotlight}.

\paragraph{Multimodal evaluation and benchmark validity.}
Multimodal benchmarks \citep{yue2024mmmu, wang2024mmlu, lu2024mathvista,
chen2024we, zuo2025medxpertqa, jiang2024mantis} score each item under
one canonical ordering. MMBench
\citep{liu2024mmbenchmultimodalmodelallaround} mitigates MCQ positional
bias via CircularEval; we generalize to five orthogonal facets and
decompose the residual via \odi. Validity work \citep{bean2026measuring,
salaudeen2025measurement, reuel2024betterbench,
jacovi2023stopuploadingtestdata}, including NIST AI 800-2
\citep{nist800-2}, treats benchmarks as instruments whose perturbation
envelope must be qualified. IRT frameworks
\citep{uebayashi2026evaluating, polo2024tinybenchmarks,
romanou2026brittlebench} target cross-modal shortcuts or item
difficulty; \odi targets ordering variance via per-item $\sigma_\pi$ and
per-facet $|\delta|$.

\paragraph{Inference-time mitigation.} Self-consistency \citep{wang2023selfconsistencyimproveschainthought} is
the canonical training-free mitigation and our baseline. Permutation
self-consistency \citep{tang2024middlepermutationselfconsistencyimproves}
extends this to listwise perturbations and runs inside our
mixed-modality LLM judge (\Cref{sec:methods}). Position-debiasing
methods include PINE \citep{wang2025eliminating}, attention calibration
\citep{hsieh2024middlecalibratingpositionalattention}, multimodal
causal-bidirectional interpolation \citep{tian2025identifying}, and
swap-and-vote LLM-as-judge \citep{zheng2023judging}. \Cref{sec:mitigation} evaluates these against cross-ordering flip on the 2026 multimodal frontier.

%% file: sections/3-methods.tex
\section{Methods}
\label{sec:methods}

\paragraph{Models.}
We audit \fpNModels{} multimodal large language models served between
May 4 and May 25, 2026.
\emph{Frontier closed-source (6):}
Gemini-Pro 3.1 (\texttt{gemini-3.1-pro-preview}) and Gemini-Flash 3
(\texttt{gemini-3-flash-preview}) \citep{pichai2025new}; Claude Opus 4.7
(\texttt{claude-opus-4-7}) and Claude Sonnet 4.6
(\texttt{claude-sonnet-4-6}) \citep{anthropic2026claude47}; ChatGPT 5.5
(\texttt{gpt-5-5}) and ChatGPT 5.4-mini (\texttt{gpt-5-4-mini})
\citep{singh2025openai}.
\emph{Open-weight (12):} Qwen3.5-VL (0.8B, 2B, 4B, 9B, 27B)
\citep{qwen3.5}, InternVL3.5 (4B, 8B, 14B, 38B)
\citep{wang2025internvl3_5}, Kimi-VL-A3B-Instruct
\citep{kimiteam2025kimivltechnicalreport}, and MedGemma (4B-IT, 27B-IT)
\citep{sellergren2025medgemma}.
Closed-source inference uses each provider's API at temperature 0
where accepted (omitted on Anthropic and OpenAI reasoning APIs, which
reject the parameter); top-$p$ = 1 on Gemini; max output tokens 8192
(Anthropic, OpenAI) or 24576 (Gemini); thinking budget 8192 tokens
(Anthropic, Gemini) or reasoning-effort \texttt{medium} (OpenAI).
A common prompt family is used across models, with facet-specific
answer-format instructions (full prompt and per-model adapter details
in \Cref{sec:supp_datasets}); briefly, prompts instruct the model to
read every evidence item before answering, and closed-choice prompts
require exactly one option label preceded by ``Answer: ''. Open-weight
models are greedy-decoded in HuggingFace
\texttt{transformers} ($\geq$4.57) on a single A6000 GPU, with
\texttt{max\_new\_tokens} 512 (default) or 2048 when emitting
reasoning traces. Models load in \texttt{bfloat16} except the
InternVL3.5 family (4B/8B/14B/38B) and the $\geq$27B variants
(Qwen3.5-VL 27B, MedGemma 27B-IT), which use 4-bit
\texttt{bitsandbytes} NF4 (double quantization, \texttt{bfloat16}
compute) with FlashAttention-2.

\emph{Caveats.} The thinking/reasoning parameter has different
per-provider semantics; we use the closest cross-provider analog
above and vary Gemini's budget in the ablation
(\Cref{sec:mit_think_budget}). The May 4--25, 2026 access window
pairs with each provider ID as the reproducibility anchor.

\paragraph{Datasets and sampling.}
\fp covers \fpNDatasets{} datasets across four task families. \emph{MCQ
reasoning} (\optionorder{}): MMLU-Pro \citep{wang2024mmlu},
CommonsenseQA
\citep{talmor2019commonsenseqaquestionansweringchallenge}, and
MathVision \citep{wang2024measuring}.
\emph{Multi-hop QA and RAG} (\evchunkorder{} and \docrankorder{}):
HotpotQA \citep{yang2018hotpotqa}, MuSiQue
\citep{trivedi2022musiquemultihopquestionssinglehop}, MultiHop-RAG
\citep{tang2024multihoprag}; MedXpertQA's~\citep{zuo2025medxpertqa}
retrieved-evidence variant
also stresses \evchunkorder{}.
\emph{Multi-image VLM} (\imagesetorder{}): Mantis-Eval
\citep{jiang2024mantis}, MedFrameQA
\citep{yu2025medframeqamultiimagemedicalvqa}.
\emph{Free-form mixed-modality RAG} (\mmorder{}): MRAMG-Recipe
\citep{yu2025mramg}, MMDocRAG \citep{dong2025mmdocrag}, MultiModalQA
\citep{talmor2021multimodalqacomplexquestionanswering}.

Within each dataset we sample a fixed item subset by deterministic
uniform shuffling with seed 42, then apply dataset-specific
filtering to admit only items with the required structural support
for the target facet. Per-dataset $N$ (after filtering) ranges
70--200, with mixed-modality datasets totaling $N{=}597$ across 3
datasets; full filter list and per-dataset $N$ in
\Cref{sec:supp_datasets}. We attempt the same item set for all
\fpNModels{} models within each (facet, dataset) cell, with retrieved
evidence (RAG and multi-hop) precomputed once and replayed verbatim
across models; derived analyses report parse or drop rates when a
model has unusable outputs. We do not
filter to gold-evidence-containing items: HotpotQA and MuSiQue use
their distractor-augmented splits, so the gold answer may or may not
be derivable from the visible evidence, mirroring open-retrieval
deployment. Demoted dialog-turn and few-shot facets and their datasets
are reported in \Cref{sec:supp_facet_extras}.

\paragraph{Ordering facets.}
\fp audits five ordering facets, each defined by the unit of
permutation and the scoring rule.
\emph{\optionorder} permutes the mapping between option content and
displayed label (the option contents themselves are unchanged; only
which content gets which letter is reordered). Scoring maps the
model's predicted letter back to the source option index via the
inverse permutation, so a flip denotes a change in selected
\emph{option content}, not a change in letter; flip on letters is
uninformative because the gold letter moves with the permutation.
Per item, the number of options permuted equals the dataset's native
choice count (CSQA fixed at 5; MMLU-Pro 4--10, median 5; MathVision
4--5).
\emph{\evchunkorder} permutes the order of evidence passages within a
single flat-list context (HotpotQA, MuSiQue, MedXpertQA retrieved
variant), where each unit is a paragraph or short passage; 3--6
chunks per item.
\emph{\docrankorder} permutes the order of \emph{ranked retrieved
documents} in a RAG prompt (MultiHop-RAG, HotpotQA-replay), where
each unit is a complete document at a specific retrieval rank; 2--6
documents per item. The distinction from \evchunkorder{} is the unit
of permutation: chunks are sub-document passages presented as a flat
list, documents are complete retrieval units presented as a ranked
list.
\emph{\imagesetorder} permutes the order of multi-image inputs to a
VLM (Mantis-Eval 3--6 images, MedFrameQA 2--5 images). We apply a
position-reference screen for image-set analyses: items whose question
or answer options explicitly refer to image position are excluded from
clean image-set summaries (52/70 Mantis-Eval items excluded; 5/200
MedFrameQA items excluded).
\emph{\mmorder} permutes the entire heterogeneous text-and-image
component sequence per item in free-form RAG; outcomes are scored by
an LLM judge (below) as Bernoulli gold-match labels, paralleling the
MCQ facets while adding a separate measurement layer.

\paragraph{Permutation grammar.}
For each (item, facet), we sample $K = 6$ orderings from the facet's
permutation grammar: uniform sampling without replacement from the
$n!$ permutations of the $n$ components when $n! \geq 6$; when fewer
than six unique permutations exist, we use all unique permutations and
repeat as needed to keep six calls. $K = 6$ is set by a $K$-ablation on
MedXpertQA: $K=3$
recovers most, but not all, of the $K=6$ detectable flips at half the
inference cost, so $K = 6$ is our operating point across the panel.
Permutation manifests are fixed before inference and replayed across all
\fpNModels{} models; the planned release includes these indices so
faithful regeneration does not depend on process hash seeding.

\paragraph{Flip-rate definition.}
For model $m$ on item $i$ under facet $f$, let
$y_{i,m,f,k} \in \mathcal{Y}_f$ be the normalized answer under
ordering $k$ (source-option index for \optionorder{} and
\imagesetorder{}, normalized short-answer string for \evchunkorder{}
and \docrankorder{}, LLM-judge verdict for \mmorder{};
\Cref{sec:supp_llm_judge}). We define an
item-level any-flip indicator
\begin{equation*}
  \operatorname{flip}_{i,m,f}
  = \mathbf{1}\!\left[\,
    \bigl|\{ y_{i,m,f,k} : k = 1, \ldots, K \}\bigr| > 1
  \,\right],
\end{equation*}
and report the panel flip rate
\begin{equation*}
  \operatorname{FlipRate}_{m,f}
  = \frac{1}{N_f} \sum_{i=1}^{N_f} \operatorname{flip}_{i,m,f}.
\end{equation*}
We use item-level any-flip rather than pairwise disagreement because
it answers the deployment-natural question (``can the answer change
under permutation?''); it is $K$-monotone (larger $K$ can only
discover more disagreements). Our $K$-ablation shows that $K=3$
recovers most, but not all, of the $K=6$ detectable-flip signal,
with $K=6$ as the operating point for the panel.

\paragraph{Same-ordering control.}
To estimate ordering excess over decoder-stochastic flips, we
run the canonical ordering $K = 6$ times (same input, independent API
calls or decoder seeds) and compute a within-canonical flip rate
$\operatorname{FlipRate}_{\text{same}}$ as an upper bound on decoder
noise. We then report
\begin{equation*}
  \Delta_{\text{order}}
  = \operatorname{FlipRate}_{\text{perm}}
  - \operatorname{FlipRate}_{\text{same}}.
\end{equation*}
We interpret $\Delta_{\text{order}}$ as an estimate of excess
instability attributable to ordering, not as a literal decomposition
of individual flips; the subtraction can be negative under
decoder-noise-dominated cells and does not propagate per-cell
uncertainty. The temperature-0 decomposition is scoped to a 12-cell
panel on Gemini-Pro 3.1 and Gemini-Flash 3 (6 \optionorder{} + 4
\imagesetorder{} + 2 \evchunkorder{} cells); we report 8 verified
cells here (4 \imagesetorder{} plus MMLU-Pro \optionorder{} and
MedXpertQA \evchunkorder{} on both models). Generalization beyond
these cells, especially at deployment temperatures, is a stated
limitation (\Cref{sec:limitations}).

\paragraph{LLM-judge scoring for free-form output.}
The non-\mmorder{} facets use deterministic scoring:
\optionorder{} and \imagesetorder{} map predicted labels back to
source content under the inverse permutation, while \evchunkorder{}
and \docrankorder{} use exact-match against gold short answers.
\mmorder{} yields paragraph-form output on 2 of 3 benchmarks
(MRAMG-Recipe, MMDocRAG); MultiModalQA gold is short factoid. For all
three, we define a Bernoulli correctness outcome via an LLM judge
(Gemini Pro as primary judge, ChatGPT 5.4-mini as cross-vendor check)
that scores the $K{=}6$ item outputs with a structured equivalence
prompt. Text-flip (raw paraphrase variation) is reported
alongside as an upper bound that the judge correction collapses by
10--90 pp depending on model and benchmark. Full prompt, inter-judge
Cohen's $\kappa$, and a judge-free MMQA gold-anchor cross-check are
in \Cref{sec:supp_llm_judge} and \Cref{sec:supp_mixmod}. The
\mmorder{} outcome enters \odi (below) as a Bernoulli gold-match
label, while the judge-based measurement layer requires the
caveat in
\Cref{tab:irt_facet_decomp}.

\paragraph{\odi.}
\odi (Ordering-Decomposed Item-Response Theory) is a 2PL hierarchical
Bayesian item-response model
\citep{birnbaum1968some, embretson2025item}. For model $m$, item
$i$, facet $f$, and permutation index $o \in \{1, \ldots, K\}$, the
binary outcome $Y_{m,i,f,o} \in \{0, 1\}$ is correctness under that
ordering (deterministic scoring for non-\mmorder{} facets;
LLM-judge gold-match for \mmorder{}). The correctness logit is
\begin{align*}
  \operatorname{logit}\,p_{m,i,f,o}
    &= \alpha_i \bigl(
      \theta_m - \beta_i
      - \delta_{f,\, d(i),\, o}
      - \gamma_{i,\, o}
    \bigr), \\
  \gamma_{i,\, o} &\sim \mathcal{N}\bigl(0,\, \sigma_{\pi, i}^2\bigr), \\
  \log \sigma_{\pi, i}
    &\sim \mathcal{N}\bigl(\mu_{f(i)},\, \tau_{f(i)}^2\bigr),
\end{align*}
where $\alpha_i$ is item discrimination, $\beta_i$ is item difficulty,
$\theta_m$ is model ability, $d(i)$ is the dataset containing item
$i$, and $\delta_{f, d, o}$ is a per-(facet, dataset, permutation-index)
systematic latent offset hierarchically pooled within facet via
$\sigma_{\delta, f}$. The item-level random effect $\gamma_{i, o}$ has
variance $\sigma_{\pi, i}^2$, also hierarchically pooled within facet
through a log-Normal hyper-prior. We report per-facet posterior
medians of the item-pooled $\sigma_{\pi, i}$ and the facet-pooled
$|\delta_{f, d, o}|$ as facet-level summaries in
\Cref{tab:irt_facet_decomp}.

\emph{Indexing note and limitations.} Here $o$ is the
\emph{permutation index} over the $K = 6$ orderings sampled per item,
not a slot-position; $\delta_{f, d, o}$ therefore captures average
outcome shift across permutation indices within a (facet, dataset)
cell. Content-vs-position decomposition (whether a specific option
content or image is more salient than a specific slot) is absorbed
into $\beta_i$ and $\alpha_i$ rather than modeled separately, and the
model has no model-level random effects beyond $\theta_m$. These are
deliberate simplifications: \odi is a summary instrument for
ordering-associated variance, not a content-saliency decomposition.

Inference uses NUTS \citep{hoffman2014no} in NumPyro / JAX
\citep{bradbury2018jax, phan2019composable} (4 chains $\times$ 3000
tune + 1500 draws, target-accept 0.99, fp64). We fit two outcome
variants: \emph{modal-outcome} ($Y = 1$ iff the answer under ordering
$o$ matches the model's modal answer for that item, emphasizing
ordering-associated instability rather than item difficulty) for the facet-level
$\sigma_\pi$ and $|\delta|$ summaries; \emph{correct-outcome}
($Y = 1$ iff the answer matches gold) for $\theta_{\text{correct}}$
in \hyperref[fig:teaser]{Figure~\ref{fig:teaser}}. Priors, NUTS
settings, $\hat R$, ESS, and divergence diagnostics are in
\Cref{sec:supp_irt_methodology}.

%% file: sections/4-findings.tex
\section{Findings}
\label{sec:findings}


\begin{figure*}[t!]
    \centering
    \includegraphics[width=1\textwidth]{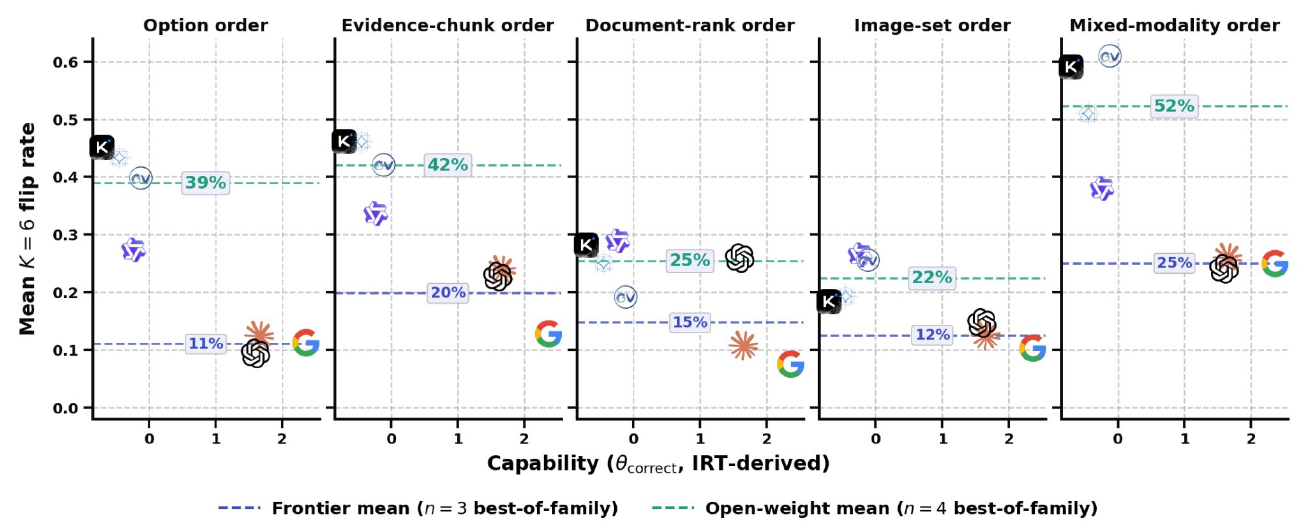}
    \caption{\textbf{Per-facet ordering sensitivity.}
    Capability ($\theta_{\text{correct}}$, IRT-derived) vs per-facet
    $K{=}6$ flip rate across the 5 facets, for $n{=}7$ best-of-family
    models (vendor-logo markers): Gemini-Pro 3.1, Claude Opus 4.7,
    ChatGPT 5.5, Qwen3.5-VL 27B, InternVL3.5 14B, Kimi-VL-A3B-Instruct,
    MedGemma 27B-IT. Dashed references mark frontier ($n{=}3$) and
    open-weight ($n{=}4$) best-of-family means. \mmorder{} uses LLM-judge
    sem-flip (\Cref{sec:supp_mixmod}). Mixed-modality is highest,
    doc-rank is lowest, and image-set is reported after the
    \mantiseval{} position-reference screen.}
    \label{fig:per_facet_overview}
\end{figure*}

\begin{table*}[t!]
    \centering
    \caption{\textbf{Per-facet flip rate by model.}
    Mean $K{=}6$ flip across 5 facets, plus a 5-facet mean column.
    Each facet column is a cell-equal mean over its balanced main-panel
    datasets; the 5-facet mean averages the facet columns equally.
    Per-column best \textbf{bold},
    second-best \underline{underlined}. \mmorder{} is LLM-judge
    sem-flip averaged across 3 mm datasets (MMDocRAG, MMQA, MRAMG).
    \imagesetorder{} uses the clean position-reference-screened subset.
    Aggregation details in \Cref{sec:supp_per_facet_tables}; capability
    in \hyperref[fig:teaser]{Figure~\ref{fig:teaser}}.}
    \label{tab:per_facet_per_model}
    \adjustbox{max width=\textwidth}{
    \begin{tabular}{lcccccc}
        \toprule[1.5pt]
        Model & \optionorder & \evchunkorder & \docrankorder & \imagesetorder & \mmorder & Mean (5-facet) \\
        \midrule
        Claude Opus 4.7        & 0.13          & 0.24          & \underline{0.11}   & \underline{0.13}   & 0.26          & \underline{0.17}   \\
        Claude Sonnet 4.6      & 0.20          & 0.28          & 0.33          & 0.21          & 0.30          & 0.26          \\
        ChatGPT 5.5            & \textbf{0.09} & 0.23          & 0.26          & 0.15          & \textbf{0.24} & 0.19          \\
        ChatGPT 5.4-mini       & 0.23          & 0.40          & 0.27          & 0.31          & 0.36          & 0.31          \\
        Gemini-Pro 3.1         & \underline{0.11}   & \textbf{0.13} & \textbf{0.08} & \textbf{0.10} & \underline{0.25}   & \textbf{0.13} \\
        Gemini-Flash 3         & 0.17          & \underline{0.20}   & 0.13          & 0.14          & 0.26          & 0.18          \\
        \midrule
        Qwen3.5-VL 0.8B        & 0.79          & 0.72          & 0.41          & 0.35          & 0.84          & 0.62          \\
        Qwen3.5-VL 2B          & 0.67          & 0.57          & 0.41          & 0.58          & 0.71          & 0.59          \\
        Qwen3.5-VL 4B          & 0.46          & 0.47          & 0.23          & 0.33          & 0.53          & 0.40          \\
        Qwen3.5-VL 9B          & 0.40          & 0.41          & 0.28          & 0.20          & 0.52          & 0.36          \\
        Qwen3.5-VL 27B         & 0.27          & 0.34          & 0.29          & 0.26          & 0.38          & 0.31          \\
        \midrule
        InternVL3.5 4B         & 0.45          & 0.50          & 0.28          & 0.23          & 0.65          & 0.42          \\
        InternVL3.5 8B         & 0.44          & 0.46          & 0.19          & 0.33          & 0.57          & 0.40          \\
        InternVL3.5 14B        & 0.40          & 0.42          & 0.19          & 0.26          & 0.61          & 0.38          \\
        InternVL3.5 38B        & 0.29          & 0.42          & 0.37          & 0.25          & 0.60          & 0.39          \\
        \midrule
        Kimi-VL-A3B-Instruct   & 0.45          & 0.46          & 0.28          & 0.18          & 0.59          & 0.39          \\
        MedGemma 4B-IT         & 0.57          & 0.58          & 0.34          & 0.19          & 0.77          & 0.49          \\
        MedGemma 27B-IT        & 0.43          & 0.46          & 0.25          & 0.19          & 0.51          & 0.37          \\
        \midrule
        \emph{Panel mean (n=18)} & \emph{0.36}   & \emph{0.41}   & \emph{0.26}   & \emph{0.24}        & \emph{0.50}   & \emph{0.35}   \\
        \bottomrule[1.5pt]
    \end{tabular}}
\end{table*}


\subsection{Cross-Facet Ordering Sensitivity}
\label{sec:cross_facet_panel}

\paragraph{Q1: How prevalent is cross-ordering flipping across the
\fpNModels-model panel?}
Cross-ordering flipping is widespread across models and facets
(\Cref{tab:per_facet_per_model}, \Cref{fig:per_facet_overview}):
per-facet panel-mean $K{=}6$ flip spans 0.24--0.50
(\mmorder{} 0.50, \evchunkorder{} 0.41, \optionorder{} 0.36,
\docrankorder{} 0.26, screened \imagesetorder{} 0.24);
\mmorder{} sem-flip is 0.50 median across 54
(model$\times$benchmark) cells, with raw \mmorder{} benchmark cells
spanning 0.09--0.89.

\paragraph{Q2: How much exceeds the same-input decoder-noise floor?}
At temperature~0, verified Gemini cells show a substantial
ordering excess over a same-input decoder-noise floor, but the share is substrate- and
temperature-conditional. Across 8 verified cells of the 12-cell panel
(\Cref{sec:supp_robustness}), the MMLU-Pro \optionorder{} cells show
an ordering-excess share of about 51--75\%. On the clean MedFrameQA
\imagesetorder{} subset, the same-ordering floor remains below the
cross-ordering rate (0.03/0.10 vs 0.10/0.14 on Pro/Flash), while
MedXpertQA \evchunkorder{} shows a similar but smaller excess
(0.11/0.15 vs 0.20/0.21). Mantis-Eval is excluded from the clean
image-set decomposition because most items contain position-reference
language. At $T \geq 0.7$, a broader appendix sweep shows the
consistency-share estimator is no longer a stable decomposition target,
whereas decoder-cleaned accuracy swing remains informative
(\Cref{sec:supp_robustness}).

\paragraph{Q3: Where are the null effects?}
We observe two reliable nulls (tool-description ordering under
disambiguating queries, and few-shot math on GSM8K), both detailed
in \Cref{sec:supp_facet_extras}.

\paragraph{Q4: Which facets carry per-item ordering noise, and which carry systematic bias?}
\label{sec:irt}
Ordering noise varies by substrate, with mixed-modality carrying the
largest per-item ordering variance in the screened modal-outcome fit.
The \odi decomposition (\Cref{sec:methods}) separates per-item
ordering noise ($\sigma_\pi$) from per-facet systematic bias
($|\delta|$) on the screened modal-outcome fit
(\Cref{tab:irt_facet_decomp}):
\imagesetorder{} is elevated in that modal-outcome fit
($\sigma_\pi$ \fpRatioSigmaImageOpt$\times$, $|\delta|$
\fpRatioDeltaImageOpt$\times$ the \optionorder{} baseline), but the
clean image-set interpretation is anchored on \medframeqa{} after the
position-reference screen. \mmorder{} has
the largest $\sigma_\pi$ overall ($3.03\times$); its large $|\delta|$
is a latent IRT offset estimated from LLM-judge gold-match labels, so
the judge mechanism may add measurement variance and inflate the
magnitude (\Cref{tab:irt_facet_decomp}, $\dagger$). Posterior
medians, 89\% HDIs, cross-facet KS separability, and per-parameter
diagnostics are in \Cref{sec:supp_irt_methodology}. A companion correct-outcome fit supplies $\theta_{\text{correct}}$ for the capability analysis in \hyperref[fig:teaser]{Figure~\ref{fig:teaser}} and this section.

\begin{table}[h!]
    \centering
    \caption{\textbf{Per-facet ordering noise and systematic bias.}
    Posterior median $\sigma_\pi$ and $|\delta|$ from the screened modal-outcome
    \odi fit; ratios vs \optionorder{} in parentheses. Bold marks the
    largest non-\mmorder{} facet; \mmorder{} is shown separately because its
    LLM-judge outcome mechanism adds measurement variance. The
    \imagesetorder{} row should be read with the position-reference
    screen caveat in \Cref{sec:limitations}. $\dagger$:
    \mmorder{} outcomes are LLM-judge gold-match Bernoulli labels;
    because $|\delta|$ is a latent IRT offset estimated from those
    labels, the judge mechanism may add measurement variance and
    inflate the value.}
    \label{tab:irt_facet_decomp}
    \adjustbox{max width=\columnwidth}{
    \begin{tabular}{lcc}
        \toprule[1.5pt]
        Facet & $\sigma_\pi$ (ratio) & $|\delta|$ (ratio) \\
        \midrule
        \optionorder      & $0.088~(1.00\times)$               & $0.003~(1.00\times)$ \\
        \docrankorder     & $0.097~(1.10\times)$               & $0.010~(3.81\times)$ \\
        \evchunkorder     & $0.105~(1.20\times)$               & $0.012~(4.50\times)$ \\
        \imagesetorder    & $\mathbf{0.152}~(\mathbf{1.73\times})$ & $\mathbf{0.022}~(\mathbf{8.25\times})$ \\
        \mmorder          & $0.265~(3.03\times)$               & $4.796^{\dagger}$ \\
        \bottomrule[1.5pt]
    \end{tabular}}
\end{table}


\subsection{Capability, Calibration, and Mechanism}
\label{sec:calibration_and_decoupling}

\paragraph{Q5: Does capability scale with ordering robustness?}
Capability predicts flips but does not eliminate them.
Panel-wide, capability and ordering robustness are tightly coupled:
Spearman $\rho(\theta_{\text{correct}}, \text{mean } K{=}6
\text{ flip}) \approx -0.95$ across the \fpNModels-model panel
(screened 5-facet flip means with screened $\theta_{\text{correct}}$)
(\hyperref[fig:teaser]{Figure~\ref{fig:teaser}}(b)); best-of-family
cluster means are \fpFrontierSubsetMean{} (n=3 frontier) vs
\fpOpenWeightSubsetMean{} (n=4 open-weight). The 18 models are not
independent samples (families share training-data lineages,
architectures, and vendor-specific serving behaviors), and
$\theta_{\text{correct}}$ is recovered from the correct-outcome
\odi fit on the same trial outcomes that produce flip rate; we
therefore read this correlation as a descriptive cross-panel summary
rather than a generalizable law. Scaling helps but does not eliminate:
the best model (Pro) still flips on \fpBestFlip{}. Within open-weight
families, parameter count reduces screened 4-facet flip monotonically
up to mid-scale (Qwen3.5-VL 0.8B$\to$27B: $0.57 \to 0.29$;
InternVL3.5 4B$\to$14B: $0.37 \to 0.32$, slight 38B uptick to 0.33;
\Cref{fig:supp_within_family_scaling}) but never closes the gap to the
frontier mean. Within the frontier cluster, capability rank does not fully
determine robustness rank (within-frontier Spearman $\rho \approx -0.89$, n=6,
qualitative): Opus over-performs its capability rank while Sonnet
under-performs (\Cref{tab:per_facet_per_model}; per-model $\theta$
and flip rates in \Cref{sec:supp_irt_methodology}). The spread is
mechanism-conditional (\Cref{sec:mechanism_subsec}).

\paragraph{Q6: Does stated confidence calibrate to ordering risk?}
\label{sec:calibration_subsec}
Self-reported confidence under-predicts cross-ordering flip rate
across all 7 models tested; mismatch shrinks with capability but
persists. On a MedXpertQA multi-turn elicitation (4 turns, with a
0--1 confidence after each of three evidence chunks; protocol in
\Cref{sec:supp_robustness}), we define mismatch as
$\text{flip rate} - (1 - \text{turn-3 confidence})$. Across 7 models
tested, mismatch ranges 8.8--62.1 pp (Opus $+8.8$ best to ChatGPT
5.4-mini $+62.1$ worst; Spearman $\approx -0.57$ vs
$\theta_{\text{correct}}$). In this protocol, the ordering-risk
component is not reflected in the self-reported calibration signal,
even with the same-ordering noise proxy.

\paragraph{Why ordering changes the answer.}
\label{sec:mechanism_subsec}
We sample up to $N{=}50$ cross-ordering-flipped items per source-dataset
cell and submit each to a Gemini-Pro judge classifying the failure
into one of six modes (\Cref{sec:supp_mechanism}). Content
rationalization is the dominant primary-judge label across the
non-dialog categorical cells in this sample
(\mathvision{}, \medxpertqa{}, and \mmlupro{}; 58\%, n=126). The
original \imagesetorder{}
mechanism row also labels content rationalization as modal (80\%,
n=49), but it was sampled before the \mantiseval{}
position-reference screen and is therefore treated as exploratory
rather than a clean image-set estimate. Cross-family agreement on the
3-class collapse is fair on the Landis-Koch scale (Fleiss
$\kappa = 0.30$, n=225 items, 3 judges; \Cref{sec:supp_mechanism}).
The best-supported mechanism claim is that categorical-order flips commonly
rationalize over answer or evidence content; the analogous clean
visual-substrate claim requires a screened resample.


\subsection{Mitigation Menu}
\label{sec:mitigation}

Training-free mitigation can detect or reduce some ordering
sensitivity, but does not reliably eliminate it and trades off cost,
coverage, or modality transfer. We test three categories:
\emph{detection/abstention} ($K{=}2$ disagreement screen),
\emph{aggregation} ($K$-majority voting / self-consistency), and
\emph{prompt transformation} (CTA and related prompts).

\paragraph{Detection and aggregation: $K$-policies.}
\label{sec:mit_k_aggregation}
A $K{=}2$ disagreement screen (two repeated queries, abstain or
escalate on disagreement) is the cheapest deployable policy: it is
precision-1 by definition against its own detection target, with
recall $0.47$--$0.80$ against $K{=}6$ flip on non-degenerate non-image text
cells. Clean Mantis image-set cells are too small/degenerate for a
stable recall headline. Across 26 screened cells, $K{=}2$ abstain
delivers $+0.019$ selective accuracy over $K{=}1$ at cost 2; $K{=}3$
majority matches that gain at cost 3; $K{=}6$ underperforms
($+0.011$ at cost 6; \Cref{tab:cost_pareto}). $K{=}2$ is therefore
the only non-baseline Pareto improvement among deployable policies. A practical benefit of recording multiple orderings is that
aggregation and abstention policies can be evaluated post hoc on any
completed panel; extending the $K$-policy analysis beyond Gemini mainly
requires scoring existing multi-order outputs, whereas prompt
transformations such as CTA require new inference.

\begin{table}[t!]
    \centering
    \caption{\textbf{Cost-Pareto for mitigation policies.}
    Mean accuracy across 26 screened (model $\times$ facet $\times$
    dataset) cells; $\Delta$ vs $P_1$. $P_2$ ($K{=}2$ abstain, $79.7\%$
    coverage) is the only non-baseline deployable Pareto improvement; $P_5$/$P_6$ are
    oracle bounds. $P_2$'s mean is \emph{selective accuracy} (items
    where the two calls agree); treating abstention as incorrect,
    all-items accuracy is $\approx 0.59$. We assume abstained items
    trigger escalation; the selective number is the operational
    accuracy under that assumption.}
    \label{tab:cost_pareto}
    \adjustbox{max width=\columnwidth}{
    \begin{tabular}{lccc}
        \toprule[1.5pt]
        Policy & Cost & Mean acc. & $\Delta$ vs $P_1$ \\
        \midrule
        $P_1$ single-pass               & 1 & 0.629          & --- \\
        \textbf{$P_2$ $K{=}2$ abstain}  & 2 & \textbf{0.648} & $\boldsymbol{+0.019}$ \\
        $P_3$ $K{=}3$ majority          & 3 & 0.648          & $+0.019$ \\
        $P_4$ $K{=}6$ majority          & 6 & 0.640          & $+0.011$ \\
        \midrule
        $P_5$ worst-ord. oracle (l.b.)  & 6 & 0.580          & $-0.049$ \\
        $P_6$ best-ord. oracle (u.b.)   & 6 & 0.674          & $+0.046$ \\
        \bottomrule[1.5pt]
    \end{tabular}}
\end{table}

\begin{figure}[h!]
    \centering
    \includegraphics[width=1\columnwidth]{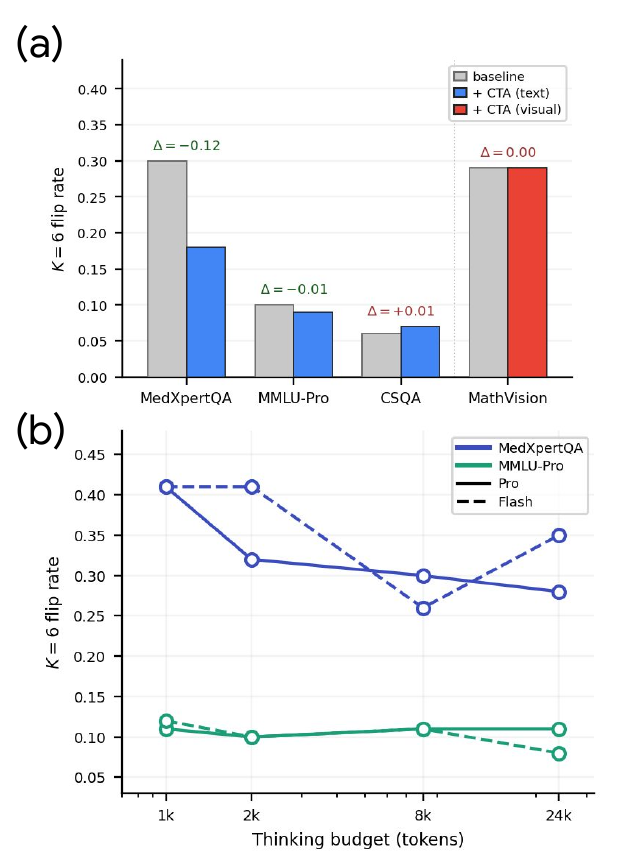}
    \caption{\textbf{Mitigation outcomes} (per-cell $n=50$ or $100$,
    $K{=}6$, $T{=}0$). \textbf{(a)}~CTA with Gemini 3.1 Pro
    at an 8{,}192-token think budget:
    $-40\%$ on the high-baseline text cell, little change on
    lower-baseline text cells,
    and no measured effect on the visual cell. \textbf{(b)}~Think-budget
    for Pro (solid) and Flash (dashed), hard \medxpertqa{}
    (\evchunkorder{}, blue) vs easy \mmlupro{} (\optionorder{}, green).
    Per-cell tables
    in \Cref{sec:supp_mitigation_extra}.}
    \label{fig:pareto}
\end{figure}

\paragraph{Prompt transformation: Canonicalize-then-Answer (CTA) is capability- and modality-asymmetric.}
\label{sec:mit_cta}
CTA prefixes the question with a model-generated canonicalization of
the evidence. On text \medxpertqa{} (Pro), CTA reduces flip $0.30 \to
0.18$ ($-40\%$); on the uncertainty-clean subset, Pro improves further
to $0.135 \to 0.029$ ($-79\%$). On Flash, MedXpertQA CTA is null to
slightly regressing ($0.26 \to 0.30$); on lower-baseline text MCQ, CTA
is marginal (Pro \mmlupro{} $0.10 \to 0.09$; \csqa{} $0.06 \to 0.07$).
On visual reasoning, CTA does not reduce flips (\mathvision{}: Pro $0.29 \to
0.29$, Flash $0.35 \to 0.35$; \Cref{fig:pareto}a) despite baselines
comparable to MedXpertQA, consistent with, but not proving, a
visual-substrate account (\Cref{sec:mechanism_subsec}). Stacking CTA
with multipass reconciliation is anti-synergistic on \medxpertqa{}
(Pro $0.18 \to 0.29$, Flash $0.30 \to 0.40$): pick one prompt-level
intervention, not several. This is the main negative result for
prompt-level mitigation on multimodal LLMs.

\paragraph{Think-budget and other prompt-level knobs.}
\label{sec:mit_think_budget}
Think-budget is dataset-conditional
(\Cref{fig:pareto}b): on hard \medxpertqa{} (\evchunkorder{}) Pro flip drops
monotonically with budget ($1k\to 24k$: $0.41\to 0.28$); on \emph{easy}
\mmlupro{} (\optionorder{}) both Pro and Flash stay essentially flat
($0.08$--$0.12$) while accuracy rises into $0.93$. Order-aware
minimal-disclaimer (P15),
Canonicalizing-CoT (T4), and PINE-lite are null or marginal in mean
(per-model deltas in \Cref{sec:supp_mitigation_extra}).

%% file: sections/5-discussion.tex
\section{Discussion}
\label{sec:discussion}

None of \fpNModels{} multimodal LLMs are order-invariant: screened
per-facet panel-mean $K{=}6$ flip ranges 24--50\%. At temperature~0,
the same-ordering Gemini control estimates a substantial ordering
excess over decoder noise in verified cells, but this
decomposition is substrate- and temperature-conditional. Capability
predicts flips but does not eliminate them (panel-wide
$\rho \approx -0.95$; best model still flips \fpBestFlip{}).
Mechanism analysis shows content rationalization as a common
categorical-cell failure mode; the original image-set mechanism row is
exploratory until repeated on the screened image-set subset. CTA reduces flip from $0.30$ to $0.18$ on the high-baseline text cell (a 12 pp absolute drop, 40\% relative reduction) and has no measured effect on the comparable-baseline visual cell.

\paragraph{Order robustness as a reporting axis.}
Canonical benchmark accuracy leaves the order-robustness dimension
uncaptured: no frontier or open-weight model card we audited reports
a cross-ordering flip rate, despite an active research literature on
the failure mode \citep{pezeshkpour2024large, chen2024premise,
laban2023you}. Following \citet{mitchell2019model} on model-card
disclosure, \citet{raji2022fallacy} on undocumented brittleness, and
NIST AI 800-2's ``qualified claims'' principle \citep{nist800-2}, we
propose cross-ordering flip rate as a standard reporting axis for
multimodal LLMs, alongside accuracy. The same-ordering control is
a methodological check: it estimates the repeated-call stochasticity
floor, so reported flip rates can be interpreted relative to
same-input instability.

\paragraph{Training-time and architectural interventions are a path forward.}
Think-budget reduces flip on hard items but plateaus on easy at
linear token cost; prompt-level mitigation is modality-conditional
and does not compose. This trade-off points toward training-time,
objective, or architectural interventions
\citep{egressy2026set} over inference-time compute. One concrete
hypothesis is that pre-training or post-training mixtures
overrepresent fixed ordering conventions (canonical option letters,
retrieval ranks, or image slots); future work should test whether
permutation-augmented or contrastive ordering objectives reduce the
effect \citep{alur2025impossibility, egressy2026set}. Documenting
input ordering as a deployment configuration parameter is an interim
safeguard.

%% file: sections/6-limitations.tex
\section{Limitations}
\label{sec:limitations}

\begin{itemize}[itemsep=1pt, topsep=0pt, leftmargin=15pt]

  \item \textbf{Same-ordering control is scoped, not panel-wide.}
  The estimate of ordering excess over decoder-stochastic flips
  runs on a 12-cell Gemini subset at temperature~0 (8 verified cells
  reported; \Cref{sec:methods}); extending the control to all
  (model, facet) cells would tighten the ordering-excess claim. The
  appendix temperature sweep adds deployment-temperature evidence, but
  the open-weight extension is at $T \in \{0.7, 1.0\}$ and is therefore
  a protocol extension rather than a matched temperature-0 comparison.

  \item \textbf{$K{=}6$ is a subsample of $n!$ and the metric is
  $K$-sensitive.} For items with $n! > 6$, reported flip rates are a
  lower bound on true sensitivity at higher $K$; the item-level
  any-flip indicator is $K$-monotone (\Cref{sec:methods}), so
  cross-paper comparisons should match $K$.

  \item \textbf{Temperature-0 API calls are not strictly deterministic.}
  Providers may include nondeterministic backend components even at
  $T{=}0$; we use the same-ordering control to quantify this floor
  rather than assume it is zero.

  \item \textbf{Clean image-set inference is anchored on MedFrameQA.}
  The image-set position-reference screen excludes 52/70 Mantis-Eval
  items and 5/200 MedFrameQA items, leaving a clean image-set subset
  of 18 Mantis-Eval and 195 MedFrameQA items. We therefore avoid
  treating Mantis-heavy image-set rows as clean estimates of visual
  order robustness.

  \item \textbf{Closed-source drift and stack asymmetry.} Provider
  APIs can update silently within their version IDs; we pin the
  May 4--25, 2026 access window (\Cref{sec:methods}) as the
  reproducibility anchor for closed-source experiments. Separately,
  closed-source uses provider APIs while open-weight uses our local
  \texttt{transformers} stack, so cross-cluster comparisons mix model
  behavior with stack-level differences (within-family open-weight
  scaling in \Cref{fig:supp_within_family_scaling} partially controls
  within open-weight).

  \item \textbf{LLM-judge sem-flip on \mmorder{} carries measurement
  variance.} Mixed-modality outcomes are LLM-judge gold-match
  Bernoulli labels like the deterministic-scored facets, but the judge
  mechanism introduces variance beyond deterministic scoring
  (\Cref{tab:irt_facet_decomp}, $\dagger$); direct cross-facet
  $|\delta|$ comparisons should be read with this caveat. In the
  Gemini MMQA temperature-sweep gold-anchor check, sem-flip exceeds
  direct gold-match flip by 4.8--11.1 pp, so the judge-corrected
  statistic is not identical to a judge-free correctness flip.

  \item \textbf{Mechanism labels are diagnostic, not causal.} The six
  mechanism classes come from an LLM-judge classification with
  cross-family Fleiss $\kappa = 0.30$ (fair on the Landis-Koch scale);
  the categorical-cell rationalization result is directionally judge-stable
  but the labels do not constitute a causal account of internal model
  behavior. The original image-set mechanism sample was collected
  before the position-reference screen and is exploratory until a
  screened resample is run.

  \item \textbf{Dataset coverage is bounded.} Our \fpNDatasets{}
  datasets cover MCQ, multi-hop QA / RAG, multi-image VQA, and
  free-form mixed-modality RAG; open-ended generation, agentic
  tool-use beyond the disambiguating-tool null, multi-turn dialog
  beyond \multichallenge{}, non-English content, and in-the-wild
  user--LLM distributions
  \citep{zhao2024wildchat, chou2025visionarena} are out of scope.

  \item \textbf{Mitigation experiments are scoped to Gemini Pro and
  Flash.} The modality-asymmetric CTA result, $K{=}2$
  disagreement-flag thresholds, and think-budget regime
  characterization all run on two frontier models; Claude and ChatGPT
  replication would push the prompt-level findings to be vendor-agnostic.

  \item \textbf{Training-time interventions are not directly tested.}
  Our finding that prompt-level mitigation is modality-conditional
  and does not compose is evidence \emph{against} a sufficient
  prompt-level fix; it is not evidence \emph{for} the sufficiency of
  training-time, objective, or architectural interventions, which we
  identify as directions for future work.

\end{itemize}

%% file: supplementary/sections/1-extended_dataset_details.tex
\section{Extended Dataset Details}
\label{sec:supp_datasets}

This appendix expands the dataset coverage map referenced from
\Cref{sec:methods}: per-dataset $N$, license, and primary facet, plus
the prompt family and per-model adapter notes that the main
paper compresses to a pointer.

\paragraph{Per-dataset summary.}
\Cref{tab:supp_datasets} lists each dataset's audited $N$ (after
deterministic seed-42 uniform sampling and facet-specific filtering),
license, and the facet it primarily stresses. Per-dataset raw audited
$N$ ranges 70--200 across the 12 datasets; the clean image-set subset
after the position-reference screen is 18 Mantis-Eval and 195
MedFrameQA items. The mixed-modality benchmarks total $N{=}597$
across 3 datasets (MRAMG-Recipe 197 after image-load filter, MMDocRAG
200, MultiModalQA 200).

\begin{table}[h!]
    \centering
    \caption{\textbf{Per-dataset coverage.} Audited $N$, license, and
    primary facet; image-set rows also report the clean
    position-reference-screened $N$ in parentheses. ``Open'' denotes a
    license permitting research use
    (CC-BY, MIT, Apache-2.0, or equivalent); see the per-paper citation
    for exact terms.}
    \label{tab:supp_datasets}
    \adjustbox{max width=\columnwidth}{
    \begin{tabular}{lccl}
        \toprule[1.5pt]
        Dataset & $N$ & License & Primary facet \\
        \midrule
        \mmlupro      & 200 & MIT          & \optionorder \\
        \csqa         & 200 & MIT          & \optionorder \\
        \medxpertqa   & 150 & CC-BY-NC-4.0 & \evchunkorder \\
        \mathvision   & 190 & CC-BY-4.0    & \optionorder \\
        \hotpotqa     & 199 & CC-BY-SA-4.0 & \evchunkorder, \docrankorder \\
        \musique      & 200 & CC-BY-4.0    & \evchunkorder \\
        \multihoprag  & 171 & MIT          & \docrankorder \\
        \mantiseval   & 70 (18 clean) & Apache-2.0   & \imagesetorder \\
        \medframeqa   & 200 (195 clean) & CC-BY-NC-4.0 & \imagesetorder \\
        MRAMG-Recipe   & 197 & CC-BY-4.0    & \mmorder \\
        MMDocRAG       & 200 & Apache-2.0   & \mmorder \\
        MultiModalQA   & 200 & MIT          & \mmorder \\
        \midrule
        \gsm{} (demoted)      & 200--500 & MIT     & few-shot \\
        \humaneval{} (demoted) & 156 & MIT    & few-shot \\
        \multichallenge{} (demoted) & 100--200 & CC-BY-4.0 & dialog-turn \\
        \bottomrule[1.5pt]
    \end{tabular}}
\end{table}

\paragraph{Facet $\times$ dataset coverage.}
The 4 deterministic-scored non-\mmorder{} facets cover all
\fpNModels{} models. \Cref{tab:per_facet_per_model} uses a balanced
two-dataset aggregation for each deterministic non-\mmorder{} facet;
additional facet stress cells, such as \mathvision{} and
\medxpertqa{}, are reported in targeted analyses. \mmorder{}
covers the 3 free-form RAG benchmarks listed above. Demoted facets
(dialog-turn, few-shot) cover a subset of (model, dataset) cells; see
\Cref{sec:supp_facet_extras}.

\paragraph{Prompt family.}
A common prompt family is used across all \fpNModels{} models and all
5 main facets, with facet-specific instructions for whether the
expected output is a single MCQ letter, a short factoid, or a
paragraph-form response. The MCQ-style prompt instructs the model to
read every evidence item before answering and to output exactly one
option label preceded by ``Answer: '' on its final line, with no other
content on that line. The full prompt and per-facet variants will be
included in the planned full code release; deviations across models
are limited to the provider-required system-message structure
(e.g., Anthropic \texttt{system} field vs Gemini
\texttt{systemInstruction}).

\paragraph{Per-model adapters.}
Closed-source inference uses the official provider SDKs at the model
IDs and access window listed in \Cref{sec:methods}. Open-weight
inference uses HuggingFace \texttt{transformers} ($\geq$4.57) on a
single A6000 with \texttt{max\_new\_tokens} 512 (default) or 2048 when
emitting reasoning traces. Models load in \texttt{bfloat16} except the
InternVL3.5 family and the $\geq$27B variants (Qwen3.5-VL 27B,
MedGemma 27B-IT), which use 4-bit \texttt{bitsandbytes} NF4 with
\texttt{bfloat16} compute and FlashAttention-2. Image inputs are converted to
each provider's expected format (base64 PNG for Anthropic and Gemini,
URL or base64 for OpenAI, \texttt{PIL.Image} for HuggingFace processors)
without resizing or quality reduction.

\paragraph{Dataset licensing.}
All 12 main-facet datasets carry open licenses that permit research
use; the two NC-licensed datasets (\medxpertqa{}, \medframeqa{}) are
used for non-commercial academic evaluation, consistent with their
terms. We redistribute no dataset content; the planned full code release
will include permutation indices, model outputs, and aggregation
scripts that reproduce the audit against the upstream sources. Items that contain
identifiable patient information were filtered upstream by the dataset
creators; we add no additional content beyond permutation metadata.

%% file: supplementary/sections/2-irt_methodology.tex
\section{Full \odi Methodology}
\label{sec:supp_irt_methodology}

This appendix expands the \odi specification from \Cref{sec:methods}:
model statement, priors, inference settings, and posterior
diagnostics for the two fits whose summaries appear in
\Cref{tab:irt_facet_decomp} and \hyperref[fig:teaser]{Figure~\ref{fig:teaser}}.

\paragraph{Model specification.}
For model $m$, item $i$, facet $f(i)$, dataset $d(i)$, and permutation
index $o \in \{1, \ldots, K\}$, the correctness logit is
\begin{align*}
  \operatorname{logit}\,p_{m,i,f,o}
    &= \alpha_i \bigl(
      \theta_m - \beta_i
      - \delta_{f,\, d,\, o}
      - \gamma_{i,\, o}
    \bigr), \\
  \gamma_{i,\, o} &\sim \mathcal{N}\bigl(0,\, \sigma_{\pi, i}^2\bigr), \\
  \log \sigma_{\pi, i}
    &\sim \mathcal{N}\bigl(\mu_{f(i)},\, \tau_{f(i)}^2\bigr).
\end{align*}
The outcome $Y_{m,i,f,o} \in \{0, 1\}$ is correctness under ordering
$o$ (deterministic scoring for the 4 non-\mmorder{} facets;
LLM-judge gold-match for \mmorder).

\paragraph{Priors.}
The screened fit uses non-centered parameterizations for the item
difficulty, systematic-offset, and ordering-noise hierarchies:
$z_{\beta,i},z_{\delta,f,d,o},z_{\sigma,i}\sim\mathcal{N}(0,1)$,
$\beta_i=\mu_{\beta,f,d}+\sigma_{\beta,f,d}z_{\beta,i}$ with
$\mu_{\beta,f,d}\sim\mathcal{N}(0,1)$ and
$\sigma_{\beta,f,d}\sim\text{HalfNormal}(0.5)$;
$\alpha_i\sim\text{LogNormal}(0,0.3)$;
$\delta_{f,d,o}=\sigma_{\delta,f}z_{\delta,f,d,o}$ with
$\sigma_{\delta,f}\sim\text{HalfNormal}(0.3)$;
$\log\sigma_{\pi,i}=\mu_{f(i)}+\tau_{f(i)}z_{\sigma,i}$ with
$\mu_f\sim\mathcal{N}(-1.2,0.5)$ and
$\tau_f\sim\text{HalfNormal}(0.1)$. Model abilities use
$\theta^{\mathrm{raw}}_m\sim\mathcal{N}(0,1)$ and are centered to
mean zero for identifiability. The hierarchical log-Normal pool on
$\sigma_{\pi,i}$ within facet keeps item-level ordering-noise
estimates shrunken toward the facet-level scale.

\paragraph{Inference.}
We run NUTS \citep{hoffman2014no} in NumPyro / JAX
\citep{bradbury2018jax, phan2019composable} (fp64;
\texttt{chain\_method=parallel}; \texttt{target\_accept=0.99};
4 chains $\times$ 3000 warmup + 1500 posterior draws). Wall-clock is
12--20 hours on a CPU node for the 18-model 5-facet panel.

\paragraph{Two fits.}
We fit two outcome variants on the same trial set.
\emph{Modal-outcome:} $Y = 1$ iff the answer under ordering $o$ matches
the model's modal answer for item $i$. This emphasizes
ordering-associated instability rather than item difficulty and is the source of the per-facet
$\sigma_\pi$ and $|\delta|$ summaries in
\Cref{tab:irt_facet_decomp}.
\emph{Correct-outcome:} $Y = 1$ iff the answer matches gold. This
recovers $\theta_{m,\text{correct}}$ used in
\hyperref[fig:teaser]{Figure~\ref{fig:teaser}}(b).

\paragraph{Posterior diagnostics.}
The screened modal-outcome fit converges cleanly:
$\hat R_{\max} = 1.000$, $\text{ESS}_{\text{bulk},\min} = 898$,
$0$ divergent transitions, $n_{\text{items}} = 3{,}612$. The screened
correct-outcome fit has $0$ divergences and well-mixed paper-critical
$\theta_{\text{correct}}$ parameters
($\hat R_{\max}=1.000$, $\text{ESS}_{\text{bulk},\min}=1{,}468$), but
its global diagnostic is weaker because several dataset-difficulty
hypermeans mix slowly ($\hat R_{\max}=1.15$,
$\text{ESS}_{\text{bulk},\min}=19$). The per-facet variance and
systematic-bias parameters remain well mixed
($\sigma_\delta$ $\hat R_{\max}=1.000$,
$\text{ESS}_{\text{bulk},\min}=830$). We therefore use the
correct-outcome fit for descriptive $\theta_{\text{correct}}$ ranking
and capability correlations, while the per-facet decomposition in
\Cref{tab:irt_facet_decomp} comes from the cleaner modal-outcome fit.
Full per-parameter posterior CSVs will be included in the planned full
code release.

\paragraph{Facet-scale posterior intervals.}
\Cref{tab:supp_odi_hdi} reports compact uncertainty summaries for the
screened modal-outcome decomposition. The $\sigma_\pi$ column is the
draw-wise per-facet median of $\sigma_{\pi,i}$. The $\sigma_\delta$
column is the direct posterior hyper-scale for systematic permutation
offsets; \Cref{tab:irt_facet_decomp} separately reports
mean absolute posterior-mean $\delta_{f,d,o}$ as its point summary.

\begin{table}[h!]
    \centering
    \caption{\textbf{Posterior uncertainty for \odi facet scales.}
    Medians and 89\% highest-density intervals are computed from
    6{,}000 posterior draws of the screened modal-outcome fit.}
    \label{tab:supp_odi_hdi}
    \footnotesize
    \adjustbox{max width=\columnwidth}{
    \begin{tabular}{lcc}
        \toprule
        Facet & $\sigma_\pi$ median [89\% HDI] & $\sigma_\delta$ [89\% HDI] \\
        \midrule
        \optionorder{} & 0.086 [0.050, 0.119] & 0.012 [0.000, 0.033] \\
        \docrankorder{} & 0.093 [0.054, 0.137] & 0.027 [0.000, 0.063] \\
        \evchunkorder{} & 0.103 [0.058, 0.146] & 0.028 [0.000, 0.060] \\
        \imagesetorder{} & 0.147 [0.078, 0.225] & 0.064 [0.000, 0.143] \\
        \mmorder{} & 0.246 [0.095, 0.416] & 2.364 [1.973, 2.764] \\
        \bottomrule
    \end{tabular}
    }
\end{table}

\paragraph{Cross-facet KS distinguishability.}
Pairwise Kolmogorov--Smirnov tests on posterior-mean
$\sigma_{\pi,i}$ item estimates from the screened modal-outcome fit
show all 10 facet pairs are separated ($p < 0.001$; many underflow to
$0$ numerically). This strong separation reflects the current model's
facet-level shrinkage of $\sigma_{\pi,i}$, so we interpret it as
evidence of facet-scale separability rather than a content-level item
classifier.

\paragraph{Per-model $\theta_{\text{correct}}$.}
Per-model posterior means and 95\% credible intervals from the
correct-outcome fit will be included as a CSV in the planned full code
release, with
items sorted by descending $\theta$. The screened fit ranks Gemini Pro
$\succ$ Gemini Flash $\succ$ Opus $\succ$ ChatGPT 5.5 $\succ$
ChatGPT 5.4-mini $\succ$ Sonnet $\succ$ best open-weight; all six
closed-source frontier models remain separated from the open-weight
cluster on $\theta_{\text{correct}}$.

\paragraph{What \odi is and is not.}
2PL Bayesian IRT is well-established
\citep{birnbaum1968some, embretson2025item, polo2024tinybenchmarks};
our contribution is the particular per-item $\sigma_\pi$ /
per-facet-and-dataset $|\delta|$ decomposition applied to the ordering
audit, not the IRT machinery itself. Item-pooled $\sigma_{\pi,i}$ is
currently shrunken within facet rather than per-item-discriminating;
content-vs-position decomposition is absorbed into $\alpha_i, \beta_i$
rather than modeled separately. These are deliberate simplifications:
\odi is a summary instrument for ordering-associated variance, not a
content-saliency decomposition.

%% file: supplementary/sections/3-per_facet_per_model_tables.tex
\section{Per-Facet $\times$ Per-Model Tables}
\label{sec:supp_per_facet_tables}

This appendix clarifies how \Cref{tab:per_facet_per_model} compresses
per-(model, facet, dataset) cells to facet means and supplies the
within-family scaling figure.

\paragraph{Reading guide.}
The 5 main facets cover all \fpNModels{} models. \Cref{tab:per_facet_per_model}
uses a balanced main-panel aggregation: two datasets per deterministic
non-\mmorder{} facet and three free-form RAG benchmarks for \mmorder{}
(MRAMG-Recipe, MMDocRAG, MultiModalQA). Per-cell flip rates and the
Ordering-Stability Index (OSI: normalized entropy of cross-ordering
answers, $1$ stable, $0$ uniform) will be included as CSVs in the
planned full code release so
readers and downstream users can re-aggregate at any granularity.
The 2 demoted facets (dialog-turn, few-shot) cover the subset of
(model, dataset) cells where data was collected; see
\Cref{sec:supp_facet_extras} for the interpretive treatment.

\begin{figure}[t!]
    \centering
    \includegraphics[width=1\columnwidth]{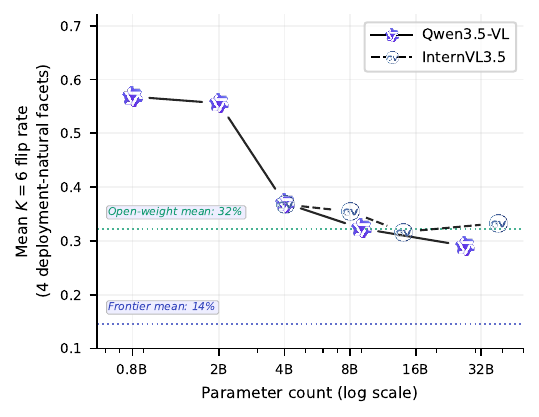}
    \caption{\textbf{Within-family scaling of ordering robustness.}
    Parameter count vs screened mean $K{=}6$ flip rate (4-facet) for Qwen3.5-VL
    and InternVL3.5 families. Scaling reduces flip monotonically within
    Qwen and within InternVL through 14B.}
    \label{fig:supp_within_family_scaling}
\end{figure}

\paragraph{Within-family scaling of ordering robustness.}
The per-(model, facet) cells in \Cref{tab:per_facet_per_model} aggregate
to per-family scaling curves on the 4 deployment-natural facets;
\Cref{fig:supp_within_family_scaling} plots these for the Qwen3.5-VL
and InternVL3.5 families. Scaling reduces 4-facet flip monotonically
within Qwen3.5-VL (0.8B$\to$27B: $0.57 \to 0.29$) and within
InternVL3.5 through 14B ($0.37 \to 0.32$, with a slight 38B uptick to
0.33), but neither family closes the gap to the frontier mean. These
screened means use the position-reference-clean image-set subset and
the screened $\theta_{\text{correct}}$ fit.

%% file: supplementary/sections/4-robustness_analysis.tex
\section{Multi-Method Robustness Analysis}
\label{sec:supp_robustness}

This appendix supports the same-ordering decomposition referenced from
\Cref{sec:cross_facet_panel} Q2 and the calibration protocol referenced
from \Cref{sec:calibration_subsec}. The methods below show that the
temperature-0 Gemini panel is not merely a decoder-noise artifact, and
then extend the decomposition at deployment temperatures with a
decoder-cleaned accuracy-swing estimator.

\paragraph{Same-ordering noise floor (12-cell decomposition).}
This is the main decomposition referenced from \Cref{sec:cross_facet_panel} Q2.
We run the canonical ordering $K{=}6$ times on Gemini-Pro 3.1 and
Gemini-Flash 3 at temperature 0 across a 12-cell panel:
6 \optionorder{} cells (\mmlupro{}, \csqa{}, \mathvision{} $\times$
Pro/Flash), 4 \imagesetorder{} cells (\mantiseval{}, \medframeqa{}
$\times$ Pro/Flash), and 2 \evchunkorder{} cells (\medxpertqa{}
retrieved variant $\times$ Pro/Flash). The main-paper interpretation
uses the position-reference-clean image-set rows: the ordering-excess
share is about 51--75\% on \mmlupro{} \optionorder{}; on the clean
\medframeqa{} \imagesetorder{} split, cross-ordering exceeds the same-ordering floor at
$T{=}0$ (floor 0.03/0.10 vs cross 0.10/0.14 on Pro/Flash) and the
floor approaches cross-ordering only at $T{\geq}0.7$. \mantiseval{} is
excluded from the clean decomposition because most items are
position-referential (\Cref{sec:limitations}). \medxpertqa{}
\evchunkorder{} likewise has a cross-ordering excess at $T{=}0$ (floor
0.11/0.15 vs cross 0.20/0.21 on Pro/Flash) and decoder-dominated
only at $T{\geq}0.7$. \multichallenge{} dialog-turn is excluded
because same-ordering accuracy ($0.007$) makes flip rate uninformative.

\paragraph{Temperature sweep and open-weight extension.}
We extend the same-ordering analysis on the clean categorical facets
(\optionorder{}, \evchunkorder{}, \docrankorder{}) with Gemini at
$T \in \{0,0.7,1.0\}$ and the open-weight families at
$T \in \{0.7,1.0\}$. This is an appendix extension rather than a
matched replacement for the temperature-0 Gemini control: the
open-weight grids have no local $T{=}0$ match. The consistency-share
estimator is masked by decoder noise at deployment temperatures: Gemini
$\Delta_{\text{order}}$ shrinks from $T{=}0$ to average $T{>}0$ on
\optionorder{} (Flash 0.049 to 0.022; Pro 0.052 to 0.017),
\evchunkorder{} (0.048 to 0.013; 0.051 to 0.013), and
\docrankorder{} (0.022 to 0.008; 0.024 to 0.008). The
decoder-cleaned accuracy-swing estimator remains more stable over the
same sweep (Flash 0.173 to 0.141 and Pro 0.134 to 0.102 on
\optionorder{}; analogous T$>$0 Gemini means are 0.100 on
\evchunkorder{} and 0.052 on \docrankorder{}). Under that estimator,
the open-weight families show a persistent robustness gap at $T{>}0$:
mean local acc-swing is 0.458 on \optionorder{}, 0.291 on
\evchunkorder{}, and 0.164 on \docrankorder{}, about 3--4$\times$ the
corresponding Gemini T$>$0 means.

\paragraph{$K{=}3$ multi-trial averaging.}
As a separate corroborating check on the 6 frontier closed-source
models, we average each (item, ordering) cell across $K{=}3$ repeated
calls. The majority-voted cross-flip is within $\pm 0.04$ of the $K{=}1$
cross-flip across all 6 closed-source models, consistent with the $K{=}1$
facet-panel data not being noise-dominated. Per-model $K{=}3$
within-cell disagreement in this diagnostic is: Opus $0.063$,
Sonnet $0.270$, ChatGPT 5.5 $0.287$, ChatGPT 5.4-mini $0.310$,
Pro $0.229$, Flash $0.188$. The stricter Gemini same-ordering
noise-floor runs give Pro $0.127$ and Flash $0.187$.

\paragraph{Within-dialog three-turn quasi-replication.}
Multi-turn elicitation forces 3 commits at the same input; the fraction
of dialogs whose 3 turns produce $\geq 2$ distinct answers is
$44.7\%$ (Opus), $49.2\%$ (Flash), $50.3\%$ (ChatGPT 5.4-mini),
$51.7\%$ (Pro), $52.0\%$ (ChatGPT 5.5), and $68.3\%$ (Sonnet). Multi-turn
re-evaluation adds variance, so within-dialog 3-turn is an upper bound
on same-ordering noise; we use it as a noise-floor proxy for the
calibration analysis (\Cref{sec:calibration_subsec}).

\paragraph{Bootstrap 95\% CIs.}
A 1000-iteration item-level bootstrap on per-model cross-ordering flip
outcomes yields CIs of approximately $\pm 8$ pp at $N{=}150$ (Pro) and
$\pm 12$--$15$ pp at $N{=}50$ (the 4 closed-source counterparts on the
add-on cells). The per-cell bootstrap CIs will be included as CSVs in
the planned full code release.

\paragraph{Stability-subset Kendall $\tau$.}
We compare per-model rankings on the full panel versus the stability
subset (items where 6/6 orderings agree). Kendall $\tau = +1.000$
($p = 0.006$) across the 6 closed-source models: rankings are exactly
preserved. The ordering signal is not driven by a borderline-item
subpopulation.

\paragraph{Random-answer permutation test.}
We pool answer letters globally, resample per item, and recompute flip
rate (1000 iterations). The observed flip rates for the 6
closed-source models are far below the pooled-shuffle null
($0.22$--$0.54$ observed versus null means near $0.99$). This is a
sanity check that models produce item-specific (non-random) answers;
it is not a significance test for the ordering effect itself.

\paragraph{Calibration protocol (\Cref{sec:calibration_subsec}).}
For Q6, we elicit confidence on \medxpertqa{} via a 4-turn protocol:
the model receives one evidence chunk at a time over 3 turns and emits
a $0$--$1$ confidence after each; the 4th turn collects the final
answer commit. Mismatch is computed as
$\text{flip rate} - (1 - \text{turn-3 confidence})$; positive mismatch
means the model is over-confident relative to its own ordering
instability.

%% file: supplementary/sections/5-llm_judge_methodology.tex
\section{LLM-Judge Methodology}
\label{sec:supp_llm_judge}

The paper uses three LLM-judge instruments: dialog-rescore for the
demoted dialog-turn facet (\Cref{sec:supp_facet_extras}), mechanism
classification for the substrate-conditional split
(\Cref{sec:mechanism_subsec}), and mixed-modality sem-flip for the
5th main facet (\Cref{sec:supp_mixmod}). Each is documented below,
followed by the shared protocol used across all three.

\paragraph{Judge 1 --- dialog-rescore.}
We pair a cross-family judge (Qwen3-VL-8B-Instruct, local) with a
cross-generation judge (Gemini 2.5 Flash Lite) and rescore
content-equivalence on dialog-turn outputs that pass raw exact-match
but differ semantically. Inter-judge Cohen's $\kappa = 0.38$ (fair on
the Landis-Koch scale); per-stratum ARI $= 0.53$. The fair-not-substantial
agreement is reported as a measurement caveat in
\Cref{sec:supp_facet_extras}.

\paragraph{Judge 2 --- mechanism classification.}
Three judges (Gemini Pro primary, ChatGPT 5.5 cross-family, Claude
Opus 4.7 cross-family) each classify the same $N{=}225$
cross-ordering-flipped items into one of six mechanism modes
(\Cref{sec:supp_mechanism} lists the taxonomy and the verbatim prompt).
Fleiss' $\kappa$ on the 3-class collapse (reasoning instability,
anchoring, other) is $0.30$ (fair). The categorical-cell finding is preserved at
all reported granularities: the modal failure is content
rationalization over answer or evidence content. The original
image-set mechanism row was sampled before the \mantiseval{}
position-reference screen and is retained only as an exploratory
diagnostic until a screened resample is run.

\paragraph{Judge 3 --- mixed-modality sem-flip.}
The primary judge is Gemini Pro with one structured equivalence prompt
over the $K{=}6$ item outputs; ChatGPT 5.4-mini serves as the
cross-vendor check.
Cross-vendor pooled 2-class (equivalent / not-equivalent) Cohen's
$\kappa$ by output-model family:
ChatGPT-output cells $\kappa = 0.81$ (substantial),
Gemini-output cells $\kappa = 0.73$ (substantial),
Claude-output cells $\kappa = 0.59$ (moderate; the lower number reflects a
prevalence paradox where Claude cells have a low not-equivalent base
rate, which inflates the marginal probability of chance agreement).
For Gemini-family mixed-modality cells, the appendix reports
cross-vendor ChatGPT/Claude agreement checks rather than relying on the
same model family as the sole validation signal.

\paragraph{Judge-free gold-anchor cross-check (MMQA).}
MMQA has short-factoid gold on $n = 181$ items; we directly compute
gold-match flip and compare against sem-flip. In the Gemini MMQA
temperature sweep (Pro/Flash at $T \in \{0,0.7,1.0\}$), sem-flip
exceeds gold-anchor flip by $4.8$--$11.1$ pp in the predicted
direction. MRAMG and MMDocRAG have $\leq 13$ short-gold items each;
the MMQA $n{=}181$ subset is the calibration anchor.

\paragraph{Shared guardrails.}
All judge instruments use fixed structured prompts and constrained
label sets. The dialog-rescore judge additionally uses
permutation self-consistency with $K_{sc}=3$ over shuffled response
labels; mixed-modality sem-flip and mechanism classification instead
use one structured prompt per item and rely on cross-vendor checks for
validation. Parser checks and pre-flight sanity probes are applied where
the output format requires clustering or structured JSON.

\paragraph{Limitations.}
The dialog-rescore $\kappa = 0.38$ is fair, not substantial;
human-judge calibration on a 100-item subset is future work. The
mixed-modality and mechanism checks use cross-vendor secondary judges
rather than a single-judge protocol, but human-rating validation is also
future work for all three uses.

%% file: supplementary/sections/6-additional_facet_results.tex
\section{Additional Facet Results}
\label{sec:supp_facet_extras}

This appendix houses (i) the two demoted facets (dialog-turn, few-shot)
that did not earn a main-paper row but carry useful signal, and (ii)
the tool-description null referenced from \Cref{sec:cross_facet_panel}
Q3.

\paragraph{Dialog-turn order (demoted).}
\multichallenge{} same-ordering accuracy is $0.007$, a degenerate floor
that makes flip rate uninterpretable, so the same-ordering decomposition
cannot run. The synthetic dialog benchmarks (\texttt{musique\_synth},
\texttt{hotpotqa\_synth}) show modest but nonzero exact-match flip on
Pro and Flash ($0.085$--$0.185$, equivalently $0.815$--$0.915$
stability) with macro-accuracy $0.69$--$0.95$. Separately, LLM-judge
rescoring on \textsc{MT-Eval}/\multichallenge{} dialog samples gives
content-flip rates $0.50$--$0.85$ depending on judge and model, so
semantic instability can be larger than exact-match scoring suggests.
We retain the data as a demoted result because the scoring regimes are
heterogeneous; the main-paper claim rests on the 5 retained facets.

\paragraph{Few-shot order (demoted).}
On the Gemini Pro/Flash few-shot runs, \gsm{} (math) is near-zero
($0$--$0.5\%$ flip at $n=200$), while \humaneval{} (code) is much
higher ($32$--$54\%$ on the $n=156$ filtered run; the earlier
$n=50$ diagnostic spanned $20$--$58\%$). Interpretation: math has
unique correct answers, so demonstration ordering rarely toggles
correctness for these frontier runs; code admits multiple equivalent
outputs whose preference can be demonstration-order-conditional.
Demoted because the dichotomy is a Gemini diagnostic rather than a
panel-wide claim, and few-shot prompting is increasingly outside the
deployment-default regime.

\paragraph{Tool-description ordering: a query-conditional null
(\Cref{sec:cross_facet_panel} Q3).}
Two regimes on Pro and Flash. Under $K{=}5$ disambiguating queries,
$0$ of 200 model--dataset items (1{,}200 trials) show a content flip
and gold-match accuracy is $96$--$100\%$; this is the reliable null
referenced in the main paper.
Under a harder \texttt{glaive\_k10} stress set with 6 orderings
(not restricted to the disambiguating subset),
Pro flip is $0.517$
and Flash flip is $0.550$ at gold-match accuracy $0.77$--$0.78$. The
tool-description null is therefore query-conditional, not
unconditional; the main-paper Q3 claim takes the
disambiguating-queries regime as the null.

%% file: supplementary/sections/7-mitigation_extended.tex
\section{Extended Mitigation Results}
\label{sec:supp_mitigation_extra}

This appendix expands \Cref{sec:mitigation}: full cost-Pareto data
backing \Cref{tab:cost_pareto}, per-model deltas for the prompt-level
mitigations the main paper reports in aggregate, and the
matched think-budget sweep on easy items.

\paragraph{Full cost-Pareto data.}
The 156 [cost, accuracy] policy points across 6 policies and
26 (model, facet, dataset) cells will be retained in the planned full
release artifact;
the aggregated values tabulated in \Cref{tab:cost_pareto} are means
across the 26 cells after applying the image-set screen, with $K{=}2$
abstain coverage of $79.7\%$ (the fraction of items the screen retains
rather than flagging for escalation).

\paragraph{Per-policy $\Delta$-accuracy and $\Delta$-cost.}
Versus the $K{=}1$ baseline in the screened aggregate, the deployable
policies are: $K{=}2$ abstain-on-disagree ($+0.019$ selective
accuracy, cost 2, $79.7\%$ coverage), $K{=}3$ majority ($+0.019$,
cost 3), and $K{=}6$ majority ($+0.011$, cost 6). The oracle bounds
are $-0.049$ for worst-order selection and $+0.046$ for best-order
selection. Across 26 screened cells, the disagreement-flag screen is
the only non-baseline low-cost Pareto improvement among deployable
policies.

\paragraph{Order-aware disclaimer (P15) and Canonicalizing-CoT (T4).}
Per-model $\Delta$-flip rate versus no-mitigation baseline:
on the 5 models tested (Pro, Flash, MedGemma 4B-IT, Qwen3.5-VL 8B,
InternVL3.5 8B), P15 is $+2.0$ pp on average, with Flash and MedGemma
regressing $+4$ to $+6$ pp. T4 is
null-to-negative on Pro and Flash: flip rate changes by $+2.7$ and
$+1.3$ pp, respectively, while accuracy drops by $1.7$--$1.8$ pp.
Neither beats the no-mitigation baseline on mean,
consistent with the main-paper statement that prompt-level knobs other
than CTA are null or marginal.

\paragraph{Temp-SC vs ordering-SC head-to-head.}
At matched $K \in \{1,2,3,6\}$, the two SC modes are orthogonal:
on \imagesetorder{} Temp-SC is higher by $\sim 40$ pp in the original
pre-screen, Mantis-heavy diagnostic panel (a canonical-order advantage
regime), not the clean image-set headline; on \optionorder{} Temp-SC
leads by 5--11 pp;
on \evchunkorder{} ordering-SC matches or beats Temp-SC at $K{=}6$ by
up to 9 pp. Disagreement-signature analysis suggests the two methods
capture distinct uncertainty sources, so combining them (Temp-SC across
orderings) is a candidate for future work but is more expensive than
the cost-Pareto policies we report.

\paragraph{CTA $+$ multipass anti-synergy detail.}
On \medxpertqa{}, CTA alone reduces flip to $0.180$ (Pro) and $0.300$
(Flash); stacking multipass reconciliation on top of CTA regresses
both back to $0.290$ and $0.400$ respectively ($+11$ pp on Pro,
$+10$ pp on Flash). One hypothesis is that multipass after CTA
introduces fresh permutation noise that the canonicalized context
cannot absorb. The negative result motivates the main-paper guidance
to pick one prompt-level intervention, not several.

\paragraph{Think-budget plateau detail.}
The main paper reports monotonic improvement with think-budget on
\emph{hard} \medxpertqa{} for Pro ($1k \to 24k$: $0.41 \to 0.28$). On
\emph{easy} \mmlupro{}, the later matched Pro/Flash sweep remains
essentially flat across budgets:
Pro $1k/2k/8k/24k$ flip $0.110/0.100/0.110/0.110$,
and Flash $0.120/0.100/0.110/0.080$; accuracy reaches
$0.927$--$0.938$ on Pro and $0.910$--$0.932$ on Flash.
The interpretation in the main paper holds: think-budget buys
robustness only where there is robustness headroom.

%% file: supplementary/sections/8-mixed_modality_facet.tex
\section{Mixed-Modality Facet: Extended Details}
\label{sec:supp_mixmod}

This appendix expands the 5th main facet (\mmorder), which the main
paper summarizes in \Cref{sec:methods} and \Cref{sec:cross_facet_panel}
but which needs more depth for reproducibility and for the multi-judge
and gold-anchor validation the main paper relies on.

\paragraph{Setup.}
\mmorder{} permutes the entire heterogeneous text-and-image component
sequence per item, not just one kind of content. Three free-form RAG
benchmarks: MRAMG-Recipe (paragraph-form recipe synthesis, 3--6
components per item), MMDocRAG (financial-document paragraph QA, 13--17
interleaved text/chart/table components), and MultiModalQA
(short-factoid answers with 4--15 captions + images + tables). Per-cell
$N \in \{197, 200, 200\}$, $K = 6$ orderings, $7{,}164$ trials per
model class. All \fpNModels{} models are evaluated.

\paragraph{LLM-judge sem-flip protocol.}
For each item, we submit the $K{=}6$ outputs to the Gemini-Pro judge
in one structured equivalence prompt. The judge returns an equivalent,
partial-flip, full-flip, or unparsable verdict; sem-flip is
partial-flip plus full-flip. Text-flip (the raw exact-string-mismatch
rate) is reported alongside as an upper bound that the judge correction
collapses by $10$--$90$ pp depending on model and benchmark.

\paragraph{Cross-vendor multi-judge validation.}
Pooled 2-class (equivalent / not-equivalent) Cohen's $\kappa$ for the
primary judge against a cross-vendor secondary:
ChatGPT-output cells $\kappa = 0.81$ (substantial),
Gemini-output cells $\kappa = 0.73$ (substantial),
Claude-output cells $\kappa = 0.59$ (moderate, with the prevalence-paradox
caveat noted in \Cref{sec:supp_llm_judge}).

\paragraph{Judge-free gold-anchor cross-check (MMQA).}
On the $n = 181$ MMQA items with unambiguous short-factoid gold, we
compute gold-match flip directly and compare against sem-flip. In the
Gemini MMQA temperature sweep (Pro/Flash at $T \in \{0,0.7,1.0\}$),
sem-flip exceeds gold-anchor flip by $4.8$--$11.1$ pp in the predicted
direction. MRAMG and MMDocRAG have $\leq 13$ short-gold items each; the
MMQA subset is the calibration anchor.

\paragraph{Headline findings.}
Per-cell sem-flip across the 18 frontier and open-weight cells $\times$
3 benchmarks spans $0.09$--$0.89$ (median $\sim 0.28$ on frontier,
$\sim 0.62$ on open-weight). By benchmark difficulty (frontier mean):
MRAMG $<$ MMDocRAG $<$ MMQA. Within-vendor capability ordering
(Opus $\succ$ Sonnet, Pro $\succ$ Flash, ChatGPT 5.5 $\succ$ ChatGPT 5.4-mini on
stability) replicates the pattern in the other 4 main facets. Text-flip
is $0.81$--$0.99$ across all cells (paraphrase upper bound). Per-cell
numbers will be included as CSVs in the planned full code release.

\paragraph{IRT integration on \mmorder.}
Per-trial gold-match labels (Gemini Pro primary, ChatGPT 5.4-mini
cross-check) feed the screened \odi modal fit as the 5th main-facet row,
jointly estimated with the other four main facets on the
\fpNModels-model panel. See \Cref{tab:irt_facet_decomp} for the
$\sigma_\pi$ and $|\delta|$ row.

\paragraph{Limitations specific to \mmorder.}
(i) Free-form scoring depends on LLM-judge semantic equivalence; we
mitigate via cross-vendor multi-judge plus the judge-free gold-anchor
cross-check on MMQA. (ii) The \mmorder{} outcomes are Bernoulli
gold-match labels like the deterministic-scored facets, but the
$|\delta|$ posterior is a latent IRT offset estimated from
judge-produced labels; for that reason we omit
the per-cell ratio against \optionorder{} in \Cref{tab:irt_facet_decomp}
and flag the value with $\dagger$. (iii) Human-rating validation of
the judge verdicts is future work.

%% file: supplementary/sections/9-mechanism_extended.tex
\section{Mechanism Classification: Extended Details}
\label{sec:supp_mechanism}

This appendix supports the substrate-conditional mechanism finding in
\Cref{sec:mechanism_subsec}: sampling protocol, judge prompt summary,
the 3-judge Fleiss $\kappa$ derivation, and the per-judge mode
distribution.

\paragraph{Sampling protocol.}
We sample up to $N = 50$ cross-ordering-flipped items per source-dataset
cell from the panel data, retaining 225 flips across five source-dataset
cells: two \optionorder{} cells (\mmlupro{} and \mathvision{}), one
\evchunkorder{} cell (\medxpertqa{}), one \imagesetorder{} cell
(\mantiseval{}), and one demoted dialog-turn cell
(\multichallenge{}). Per-cell item selection is uniform random over the
items with at least one detected $K = 6$ flip. The image-set
representative dataset in this original mechanism sample is Mantis-Eval;
because the position-reference screen later flags most Mantis-Eval
items, that row is an original-panel diagnostic rather than a clean
image-set estimate.

\paragraph{Judge prompt.}
The production prompt gives the judge the facet/dataset label, the
available gold answer label, and compact summaries of the six
orderings: component-order indices, normalized answer labels, and
short answer-text excerpts. The original question text was not
uniformly stored for this diagnostic pass, so the judge is asked to
classify the dominant failure pattern visible in the cross-ordering
answer summaries into one of six modes (positional anchoring, content
rationalization, hallucination, refusal shift, semantic drift, other)
and return a JSON object with the label plus a one-sentence rationale.
The full prompt text will be included in the planned full code release.

\paragraph{Per-judge classification distribution.}
Across the three judges (Gemini Pro, ChatGPT 5.5, Claude Opus 4.7),
content rationalization is the modal aggregate label on the two
\optionorder{} source cells. The main-text 58\% headline is the primary
judge's share over the three non-dialog categorical cells
(\mathvision{}, \medxpertqa{}, and \mmlupro{}; $n = 126$). On the
pre-screen \imagesetorder{} Mantis-Eval sample, content rationalization
is modal for all three judges (primary $80\%$, Claude $98\%$, ChatGPT
$55\%$ for $n = 49$), but this row is not a clean estimate after the
position-reference screen. Hallucination is most frequent on
\mathvision{} ($14\%$ in the primary judge) and is $0\%$ on the
\medxpertqa{}/\mmlupro{} text cells.

\paragraph{3-judge Fleiss $\kappa$.}
The raw 6-class taxonomy gives Fleiss $\kappa = 0.28$ (fair agreement;
judges disagree on fine-grained boundaries). The main-text 3-class
collapse (``rationalization'' vs ``anchoring'' vs ``other'') gives
$\kappa = 0.30$, matching a separate reasoning-instability collapse that
groups hallucination with rationalization. The categorical-cell finding
that flipped answers often rationalize over answer or evidence content
is preserved under this collapse. The visual analogue awaits a screened
image-set resample.

\paragraph{Aggregate mode distribution.}
On the 3-class collapse, aggregated across the original-panel $225$
items and three judges: reasoning instability (content rationalization
plus hallucination) is $59\%$ (95\% bootstrap CI $52$--$64\%$),
positional anchoring $18\%$ (CI $14$--$23\%$), and other modes $24\%$
(CI $19$--$30\%$). Bootstrap is 1000 iterations at the item level.

\paragraph{Limitations.}
(i) The taxonomy is judge-derived; we do not claim it is the final word
on flip mechanisms. (ii) $N = 50$ per cell is small relative to the
full panel; per-cell and aggregate numbers should be read as suggestive
rather than precise. (iii) The diagnostic prompt used compact answer
summaries rather than full item text, so the labels should not be read
as fine-grained causal annotations. (iv) The image-set row is pre-screen
and Mantis-heavy; clean visual-mechanism claims require a screened
resample. (v) Adding a fourth judge family (open-weight) is future work.

%% file: supplementary/sections/10-extensibility.tex
\section{Extending \fp via HuggingFace Dataset Metadata}
\label{sec:supp_extensibility}

This appendix sketches how \fp{} can be extended to new datasets and
new ordering facets with minimal engineering. The audit's per-facet
permutation grammar and scoring rule are decoupled from the underlying
dataset interface, so a new (dataset, facet) cell drops in once two
contracts are satisfied: a permutation-unit specification and a
gold-comparison rule.

\paragraph{HuggingFace-metadata-driven facet routing.}
The 12 datasets in our audit are all loaded via the
HuggingFace \texttt{datasets} library. We map each (dataset,
configuration) pair to a list of admissible facets through a small
declarative spec: the spec lists the dataset's structural support for
each facet (e.g., MMLU-Pro supports \optionorder{} because it carries
$\geq 4$ MCQ choices; MultiHop-RAG supports \docrankorder{} because it
carries ranked retrieved documents). Dataset-card YAML fields
(\texttt{task\_categories}, \texttt{modalities},
\texttt{size\_categories}) populate sensible facet defaults; the spec
overrides them where the dataset shape requires a non-default treatment
(e.g., MedXpertQA's retrieved-evidence variant is routed to
\evchunkorder{} rather than \optionorder{}).

\paragraph{Adding a new facet.}
A new ordering facet (e.g., \texttt{citation-order} on retrieval QA
with explicit citation markers) requires three components: (i) a
permutation grammar (\Cref{sec:methods}) that specifies the unit being
permuted and how to sample $K$ orderings; (ii) a scoring rule that
maps the model's raw output back to a content label invariant under
the permutation (e.g., for \optionorder{}, the inverse-permutation
mapping from displayed letter to source option index); (iii) a
same-ordering control that runs the canonical ordering $K$ times to
quantify the decoder-stochastic floor for the new facet. The first
two components slot directly into the facet registry; the third reuses the
existing same-ordering harness.

\paragraph{Adding a new dataset.}
A new dataset slots into the audit once its loader exposes the per-item
fields the target facet needs: choice list for \optionorder{}, evidence
chunks with rank metadata for \evchunkorder{} / \docrankorder{},
image list for \imagesetorder{}, and heterogeneous component sequence
for \mmorder{}. Datasets whose HuggingFace cards already declare these
fields (via \texttt{features} schema) require no adapter; datasets
without standard schemas require a $\sim 30$-line loader that maps to
the canonical per-item record. The planned full code release will
include loaders for the 12 datasets in \Cref{sec:supp_datasets} and a
template for new ones.

\paragraph{Adding a new model.}
Closed-source models slot in with a provider-API adapter (set the
endpoint, model ID, and provider-specific system-message structure).
Open-weight models slot in with a HuggingFace \texttt{transformers}
adapter (set the processor class, generation kwargs, and any
provider-specific image preprocessing). Both adapters are
$\sim 40$ lines apiece in the current implementation.

\paragraph{Scope of the release.}
The planned full code release will reproduce the audit on the 12
datasets, 5 facets, and \fpNModels{} models reported in this paper,
plus a template adapter and a registration script for adding new
datasets, facets, or models. It will include permutation indices and
aggregation scripts; upstream dataset content will be loaded from the
original HuggingFace sources rather than redistributed.

%% file: references.bib
@article{haberle2024impact,
  title={The impact of nuance DAX ambient listening AI documentation: a cohort study},
  author={Haberle, Tyler and Cleveland, Courtney and Snow, Greg L and Barber, Chris and Stookey, Nikki and Thornock, Cari and Younger, Laurie and Mullahkhel, Buzzy and Ize-Ludlow, Diego},
  journal={Journal of the American Medical Informatics Association},
  volume={31},
  number={4},
  pages={975--979},
  year={2024},
  publisher={Oxford University Press}
}

@article{tierney2024ambient,
  title={Ambient artificial intelligence scribes to alleviate the burden of clinical documentation},
  author={Tierney, Aaron A and Gayre, Gregg and Hoberman, Brian and Mattern, Britt and Ballesca, Manuel and Kipnis, Patricia and Liu, Vincent and Lee, Kristine},
  journal={NEJM Catalyst Innovations in Care Delivery},
  volume={5},
  number={3},
  pages={CAT--23},
  year={2024},
  publisher={Massachusetts Medical Society}
}

@article{hurt2025use,
  title={The use of an artificial intelligence platform OpenEvidence to augment clinical decision-making for primary care physicians},
  author={Hurt, Ryan T and Stephenson, Christopher R and Gilman, Elizabeth A and Aakre, Christopher A and Croghan, Ivana T and Mundi, Manpreet S and Ghosh, Karthik and Edakkanambeth Varayil, Jithinraj},
  journal={Journal of Primary Care \& Community Health},
  volume={16},
  pages={21501319251332215},
  year={2025},
  publisher={SAGE Publications Sage CA: Los Angeles, CA}
}

@article{lewis2020retrieval,
  title={Retrieval-augmented generation for knowledge-intensive nlp tasks},
  author={Lewis, Patrick and Perez, Ethan and Piktus, Aleksandra and Petroni, Fabio and Karpukhin, Vladimir and Goyal, Naman and K{\"u}ttler, Heinrich and Lewis, Mike and Yih, Wen-tau and Rockt{\"a}schel, Tim and others},
  journal={Advances in neural information processing systems},
  volume={33},
  pages={9459--9474},
  year={2020}
}

@article{abootorabi2025ask,
  title={Ask in any modality: A comprehensive survey on multimodal retrieval-augmented generation},
  author={Abootorabi, Mohammad Mahdi and Zobeiri, Amirhosein and Dehghani, Mahdi and Mohammadkhani, Mohammadali and Mohammadi, Bardia and Ghahroodi, Omid and Baghshah, Mahdieh Soleymani and Asgari, Ehsaneddin},
  journal={Findings of the Association for Computational Linguistics: ACL 2025},
  pages={16776--16809},
  year={2025}
}

@inproceedings{yao2025spotlight,
  title={Who is in the Spotlight: The Hidden Bias Undermining Multimodal Retrieval-Augmented Generation},
  author={Yao, Jiayu and Liu, Shenghua and Wang, Yiwei and Mei, Lingrui and Bi, Baolong and Ge, Yuyao and Li, Zhecheng and Cheng, Xueqi},
  booktitle={Proceedings of the 2025 Conference on Empirical Methods in Natural Language Processing},
  pages={15194--15204},
  year={2025}
}

@article{shinn2023reflexion,
  title={Reflexion: Language agents with verbal reinforcement learning},
  author={Shinn, Noah and Cassano, Federico and Gopinath, Ashwin and Narasimhan, Karthik and Yao, Shunyu},
  journal={Advances in neural information processing systems},
  volume={36},
  pages={8634--8652},
  year={2023}
}

@inproceedings{faghih2025tool,
  title={Tool Preferences in Agentic LLMs are Unreliable},
  author={Faghih, Kazem and Wang, Wenxiao and Cheng, Yize and Bharti, Siddhant and Sriramanan, Gaurang and Balasubramanian, Sriram and Hosseini, Parsa and Feizi, Soheil},
  booktitle={Proceedings of the 2025 Conference on Empirical Methods in Natural Language Processing},
  pages={20965--20980},
  year={2025}
}

@article{mei2025survey,
  title={A survey of context engineering for large language models},
  author={Mei, Lingrui and Yao, Jiayu and Ge, Yuyao and Wang, Yiwei and Bi, Baolong and Cai, Yujun and Liu, Jiazhi and Li, Mingyu and Li, Zhong-Zhi and Zhang, Duzhen and others},
  journal={arXiv preprint arXiv:2507.13334},
  year={2025}
}

@article{bean2026measuring,
  title={Measuring what matters: Construct validity in large language model benchmarks},
  author={Bean, Andrew M and Kearns, Ryan Othniel and Romanou, Angelika and Hafner, Franziska Sofia and Mayne, Harry and Batzner, Jan and Foroutan Eghlidi, Negar and Schmitz, Chris and Korgul, Karolina and Batra, Hunar and others},
  journal={Advances in Neural Information Processing Systems},
  volume={38},
  year={2026}
}

@article{salaudeen2025measurement,
  title={Measurement to meaning: A validity-centered framework for ai evaluation},
  author={Salaudeen, Olawale and Reuel, Anka and Ahmed, Ahmed and Bedi, Suhana and Robertson, Zachary and Sundar, Sudharsan and Domingue, Ben and Wang, Angelina and Koyejo, Sanmi},
  journal={arXiv preprint arXiv:2505.10573},
  year={2025}
}

@misc{nist800-2,
  title  = {Practices for Automated Benchmark Evaluations of Language Models},
  author = {Keller, Drew and Steed, Ryan and Wang, Tony and Bergman, Stevie and Cihon, Peter},
  year   = {2026},
  note   = {NIST AI 800-2 Initial Public Draft},
  url    = {https://nvlpubs.nist.gov/nistpubs/ai/NIST.AI.800-2.ipd.pdf}
}

@article{wang2024mmlu,
  title={Mmlu-pro: A more robust and challenging multi-task language understanding benchmark},
  author={Wang, Yubo and Ma, Xueguang and Zhang, Ge and Ni, Yuansheng and Chandra, Abhranil and Guo, Shiguang and Ren, Weiming and Arulraj, Aaran and He, Xuan and Jiang, Ziyan and others},
  journal={Advances in Neural Information Processing Systems},
  volume={37},
  pages={95266--95290},
  year={2024}
}

@article{zuo2025medxpertqa,
  title={Medxpertqa: Benchmarking expert-level medical reasoning and understanding},
  author={Zuo, Yuxin and Qu, Shang and Li, Yifei and Chen, Zhangren and Zhu, Xuekai and Hua, Ermo and Zhang, Kaiyan and Ding, Ning and Zhou, Bowen},
  journal={arXiv preprint arXiv:2501.18362},
  year={2025}
}

@article{jiang2024mantis,
  title={Mantis: Interleaved multi-image instruction tuning},
  author={Jiang, Dongfu and He, Xuan and Zeng, Huaye and Wei, Cong and Ku, Max and Liu, Qian and Chen, Wenhu},
  journal={arXiv preprint arXiv:2405.01483},
  year={2024}
}

@article{tan2024order,
  title={Order matters: Exploring order sensitivity in multimodal large language models},
  author={Tan, Zhijie and Chu, Xu and Li, Weiping and Mo, Tong},
  journal={arXiv preprint arXiv:2410.16983},
  year={2024}
}

@article{ismithdeen2025promptception,
  title={Promptception: How Sensitive Are Large Multimodal Models to Prompts?},
  author={Ismithdeen, Mohamed Insaf and Khattak, Muhammad Uzair and Khan, Salman},
  journal={arXiv preprint arXiv:2509.03986},
  year={2025}
}

@article{uebayashi2026evaluating,
  title={Evaluating Cross-Modal Reasoning Ability and Problem Characteristics with Multimodal Item Response Theory},
  author={Uebayashi, Shunki and Masui, Kento and Atarashi, Kyohei and Bao, Han and Kashima, Hisashi and Inoue, Naoto and Otani, Mayu and Takeuchi, Koh},
  journal={arXiv preprint arXiv:2603.02663},
  year={2026}
}

@article{kostiuk2026one,
  title={One prompt is not enough: Instruction Sensitivity Undermines Embedding Model Evaluation},
  author={Kostiuk, Yevhen and Enevoldsen, Kenneth},
  journal={arXiv preprint arXiv:2605.22544},
  year={2026}
}

@inproceedings{pezeshkpour2024large,
  title={Large language models sensitivity to the order of options in multiple-choice questions},
  author={Pezeshkpour, Pouya and Hruschka, Estevam},
  booktitle={Findings of the Association for Computational Linguistics: NAACL 2024},
  pages={2006--2017},
  year={2024}
}

@article{laban2023you,
  title={Are you sure? challenging llms leads to performance drops in the flipflop experiment},
  author={Laban, Philippe and Murakhovs' ka, Lidiya and Xiong, Caiming and Wu, Chien-Sheng},
  journal={arXiv preprint arXiv:2311.08596},
  year={2023}
}

@article{chen2024premise,
  title={Premise order matters in reasoning with large language models},
  author={Chen, Xinyun and Chi, Ryan A and Wang, Xuezhi and Zhou, Denny},
  journal={arXiv preprint arXiv:2402.08939},
  year={2024}
}

@article{paruchuri2025s,
  title={" what’s up, doc?": Analyzing how users seek health information in large-scale conversational ai datasets},
  author={Paruchuri, Akshay and Aziz, Maryam and Vartak, Rohit and Ali, Ayman and Uchehara, Best and Liu, Xin and Chatterjee, Ishan and Agrawal, Monica},
  journal={arXiv preprint arXiv:2506.21532},
  year={2025}
}

@inproceedings{lu2022fantastically,
  title={Fantastically ordered prompts and where to find them: Overcoming few-shot prompt order sensitivity},
  author={Lu, Yao and Bartolo, Max and Moore, Alastair and Riedel, Sebastian and Stenetorp, Pontus},
  booktitle={Proceedings of the 60th Annual Meeting of the Association for Computational Linguistics (Volume 1: Long Papers)},
  pages={8086--8098},
  year={2022}
}

@article{liu2024lost,
  title={Lost in the middle: How language models use long contexts},
  author={Liu, Nelson F and Lin, Kevin and Hewitt, John and Paranjape, Ashwin and Bevilacqua, Michele and Petroni, Fabio and Liang, Percy},
  journal={Transactions of the association for computational linguistics},
  volume={12},
  pages={157--173},
  year={2024}
}

@inproceedings{zheng2024large,
  title={Large language models are not robust multiple choice selectors},
  author={Zheng, Chujie and Zhou, Hao and Meng, Fandong and Zhou, Jie and Huang, Minlie},
  booktitle={International Conference on Learning Representations},
  volume={2024},
  pages={19426--19454},
  year={2024}
}

@inproceedings{sclar2024quantifying,
  title={Quantifying Language Models' Sensitivity to Spurious Features in Prompt Design or: How I learned to start worrying about prompt formatting},
  author={Sclar, Melanie and Choi, Yejin and Tsvetkov, Yulia and Suhr, Alane},
  booktitle={International Conference on Learning Representations},
  volume={2024},
  pages={25055--25083},
  year={2024}
}

@inproceedings{tian2025identifying,
  title={Identifying and mitigating position bias of multi-image vision-language models},
  author={Tian, Xinyu and Zou, Shu and Yang, Zhaoyuan and Zhang, Jing},
  booktitle={Proceedings of the Computer Vision and Pattern Recognition Conference},
  pages={10599--10609},
  year={2025}
}

@inproceedings{yue2024mmmu,
  title={Mmmu: A massive multi-discipline multimodal understanding and reasoning benchmark for expert agi},
  author={Yue, Xiang and Ni, Yuansheng and Zhang, Kai and Zheng, Tianyu and Liu, Ruoqi and Zhang, Ge and Stevens, Samuel and Jiang, Dongfu and Ren, Weiming and Sun, Yuxuan and others},
  booktitle={Proceedings of the IEEE/CVF conference on computer vision and pattern recognition},
  pages={9556--9567},
  year={2024}
}

@article{chen2024we,
  title={Are we on the right way for evaluating large vision-language models?},
  author={Chen, Lin and Li, Jinsong and Dong, Xiaoyi and Zhang, Pan and Zang, Yuhang and Chen, Zehui and Duan, Haodong and Wang, Jiaqi and Qiao, Yu and Lin, Dahua and others},
  journal={Advances in Neural Information Processing Systems},
  volume={37},
  pages={27056--27087},
  year={2024}
}

@article{reuel2024betterbench,
  title={Betterbench: Assessing ai benchmarks, uncovering issues, and establishing best practices},
  author={Reuel, Anka and Hardy, Amelia and Smith, Chandler and Lamparth, Max and Hardy, Malcolm and Kochenderfer, Mykel J},
  journal={Advances in Neural Information Processing Systems},
  volume={37},
  pages={21763--21813},
  year={2024}
}

@article{polo2024tinybenchmarks,
  title={tinyBenchmarks: evaluating LLMs with fewer examples},
  author={Polo, Felipe Maia and Weber, Lucas and Choshen, Leshem and Sun, Yuekai and Xu, Gongjun and Yurochkin, Mikhail},
  journal={arXiv preprint arXiv:2402.14992},
  year={2024}
}

@article{romanou2026brittlebench,
  title={Brittlebench: Quantifying LLM robustness via prompt sensitivity},
  author={Romanou, Angelika and Ibrahim, Mark and Ross, Candace and Shaib, Chantal and Oktar, Kerem and Bell, Samuel J and Ovalle, Anaelia and Dodge, Jesse and Bosselut, Antoine and Sinha, Koustuv and others},
  journal={arXiv preprint arXiv:2603.13285},
  year={2026}
}

@inproceedings{wang2025eliminating,
  title={Eliminating position bias of language models: A mechanistic approach},
  author={Wang, Ziqi and Zhang, Hanlin and Li, Xiner and Huang, Kuan-Hao and Han, Chi and Ji, Shuiwang and Kakade, Sham and Peng, Hao and Ji, Heng},
  booktitle={International Conference on Learning Representations},
  volume={2025},
  pages={91212--91239},
  year={2025}
}

@article{zheng2023judging,
  title={Judging llm-as-a-judge with mt-bench and chatbot arena},
  author={Zheng, Lianmin and Chiang, Wei-Lin and Sheng, Ying and Zhuang, Siyuan and Wu, Zhanghao and Zhuang, Yonghao and Lin, Zi and Li, Zhuohan and Li, Dacheng and Xing, Eric and others},
  journal={Advances in neural information processing systems},
  volume={36},
  pages={46595--46623},
  year={2023}
}

@article{mizrahi2024state,
  title={State of what art? a call for multi-prompt llm evaluation},
  author={Mizrahi, Moran and Kaplan, Guy and Malkin, Dan and Dror, Rotem and Shahaf, Dafna and Stanovsky, Gabriel},
  journal={Transactions of the Association for Computational Linguistics},
  volume={12},
  pages={933--949},
  year={2024},
  publisher={MIT Press 255 Main Street, 9th Floor, Cambridge, Massachusetts 02142, USA~…}
}

@article{alur2025impossibility,
  title={The Impossibility of Inverse Permutation Learning in Transformer Models},
  author={Alur, Rohan and Hays, Chris and Raghavan, Manish and Shah, Devavrat},
  journal={arXiv preprint arXiv:2509.24125},
  year={2025}
}

@misc{liu2024mmbenchmultimodalmodelallaround,
      title={MMBench: Is Your Multi-modal Model an All-around Player?}, 
      author={Yuan Liu and Haodong Duan and Yuanhan Zhang and Bo Li and Songyang Zhang and Wangbo Zhao and Yike Yuan and Jiaqi Wang and Conghui He and Ziwei Liu and Kai Chen and Dahua Lin},
      year={2024},
      eprint={2307.06281},
      archivePrefix={arXiv},
      primaryClass={cs.CV},
      url={https://arxiv.org/abs/2307.06281}, 
}

@misc{jacovi2023stopuploadingtestdata,
      title={Stop Uploading Test Data in Plain Text: Practical Strategies for Mitigating Data Contamination by Evaluation Benchmarks}, 
      author={Alon Jacovi and Avi Caciularu and Omer Goldman and Yoav Goldberg},
      year={2023},
      eprint={2305.10160},
      archivePrefix={arXiv},
      primaryClass={cs.CL},
      url={https://arxiv.org/abs/2305.10160}, 
}

@inproceedings{lu2024mathvista,
  author    = {Lu, Pan and Bansal, Hritik and Xia, Tony and Liu, Jiacheng and Li, Chunyuan and Hajishirzi, Hannaneh and Cheng, Hao and Chang, Kai-Wei and Galley, Michel and Gao, Jianfeng},
  title     = {MathVista: Evaluating Mathematical Reasoning of Foundation Models in Visual Contexts},
  booktitle={International Conference on Learning Representations (ICLR)},
  year      = {2024}
}

@misc{wang2023selfconsistencyimproveschainthought,
      title={Self-Consistency Improves Chain of Thought Reasoning in Language Models}, 
      author={Xuezhi Wang and Jason Wei and Dale Schuurmans and Quoc Le and Ed Chi and Sharan Narang and Aakanksha Chowdhery and Denny Zhou},
      year={2023},
      eprint={2203.11171},
      archivePrefix={arXiv},
      primaryClass={cs.CL},
      url={https://arxiv.org/abs/2203.11171}, 
}

@misc{tang2024middlepermutationselfconsistencyimproves,
      title={Found in the Middle: Permutation Self-Consistency Improves Listwise Ranking in Large Language Models}, 
      author={Raphael Tang and Xinyu Zhang and Xueguang Ma and Jimmy Lin and Ferhan Ture},
      year={2024},
      eprint={2310.07712},
      archivePrefix={arXiv},
      primaryClass={cs.CL},
      url={https://arxiv.org/abs/2310.07712}, 
}

@misc{hsieh2024middlecalibratingpositionalattention,
      title={Found in the Middle: Calibrating Positional Attention Bias Improves Long Context Utilization}, 
      author={Cheng-Yu Hsieh and Yung-Sung Chuang and Chun-Liang Li and Zifeng Wang and Long T. Le and Abhishek Kumar and James Glass and Alexander Ratner and Chen-Yu Lee and Ranjay Krishna and Tomas Pfister},
      year={2024},
      eprint={2406.16008},
      archivePrefix={arXiv},
      primaryClass={cs.CL},
      url={https://arxiv.org/abs/2406.16008}, 
}

@misc{talmor2019commonsenseqaquestionansweringchallenge,
      title={CommonsenseQA: A Question Answering Challenge Targeting Commonsense Knowledge}, 
      author={Alon Talmor and Jonathan Herzig and Nicholas Lourie and Jonathan Berant},
      year={2019},
      eprint={1811.00937},
      archivePrefix={arXiv},
      primaryClass={cs.CL},
      url={https://arxiv.org/abs/1811.00937}, 
}

@inproceedings{wang2024measuring,
title={Measuring Multimodal Mathematical Reasoning with MATH-Vision Dataset},
author={Ke Wang and Junting Pan and Weikang Shi and Zimu Lu and Houxing Ren and Aojun Zhou and Mingjie Zhan and Hongsheng Li},
booktitle={The Thirty-eight Conference on Neural Information Processing Systems Datasets and Benchmarks Track},
year={2024},
url={https://openreview.net/forum?id=QWTCcxMpPA}
}

@inproceedings{yang2018hotpotqa,
  title={{HotpotQA}: A Dataset for Diverse, Explainable Multi-hop Question Answering},
  author={Yang, Zhilin and Qi, Peng and Zhang, Saizheng and Bengio, Yoshua and Cohen, William W. and Salakhutdinov, Ruslan and Manning, Christopher D.},
  booktitle={Conference on Empirical Methods in Natural Language Processing ({EMNLP})},
  year={2018}
}

@misc{trivedi2022musiquemultihopquestionssinglehop,
      title={MuSiQue: Multihop Questions via Single-hop Question Composition}, 
      author={Harsh Trivedi and Niranjan Balasubramanian and Tushar Khot and Ashish Sabharwal},
      year={2022},
      eprint={2108.00573},
      archivePrefix={arXiv},
      primaryClass={cs.CL},
      url={https://arxiv.org/abs/2108.00573}, 
}

@misc{tang2024multihoprag,
      title={MultiHop-RAG: Benchmarking Retrieval-Augmented Generation for Multi-Hop Queries}, 
      author={Yixuan Tang and Yi Yang},
      year={2024},
      eprint={2401.15391},
      archivePrefix={arXiv},
      primaryClass={cs.CL}
}

@article{yu2025medframeqamultiimagemedicalvqa,
  title={MedFrameQA: A Multi-Image Medical VQA Benchmark for Clinical Reasoning}, 
  author={Yu, Suhao and Wang, Haojin and Wu, Juncheng and Xie, Cihang and Zhou, Yuyin},
  journal = {arXiv preprint arXiv:2505.16964},
  year={2025}
}

@article{yu2025mramg,
  title={MRAMG-Bench: A BeyondText Benchmark for Multimodal Retrieval-Augmented Multimodal Generation},
  author={Yu, Qinhan and Xiao, Zhiyou and Li, Binghui and Wang, Zhengren and Chen, Chong and Zhang, Wentao},
  journal={arXiv preprint arXiv:2502.04176},
  year={2025}
}

@misc{dong2025mmdocrag,
      title={Benchmarking Retrieval-Augmented Multimodal Generation for Document Question Answering}, 
      author={Kuicai Dong and Yujing Chang and Shijie Huang and Yasheng Wang and Ruiming Tang and Yong Liu},
      year={2025},
      eprint={2505.16470},
      archivePrefix={arXiv},
      primaryClass={cs.IR},
      url={https://arxiv.org/abs/2505.16470}, 
}

@misc{talmor2021multimodalqacomplexquestionanswering,
      title={MultiModalQA: Complex Question Answering over Text, Tables and Images}, 
      author={Alon Talmor and Ori Yoran and Amnon Catav and Dan Lahav and Yizhong Wang and Akari Asai and Gabriel Ilharco and Hannaneh Hajishirzi and Jonathan Berant},
      year={2021},
      eprint={2104.06039},
      archivePrefix={arXiv},
      primaryClass={cs.CL},
      url={https://arxiv.org/abs/2104.06039}, 
}

@misc{qwen3.5,
    title  = {{Qwen3.5}: Towards Native Multimodal Agents},
    author = {{Qwen Team}},
    month  = {February},
    year   = {2026},
    url    = {https://qwen.ai/blog?id=qwen3.5}
}

@article{wang2025internvl3_5,
  title={InternVL3.5: Advancing Open-Source Multimodal Models in Versatility, Reasoning, and Efficiency},
  author={Wang, Weiyun and Gao, Zhangwei and Gu, Lixin and Pu, Hengjun and Cui, Long and Wei, Xingguang and Liu, Zhaoyang and Jing, Linglin and Ye, Shenglong and Shao, Jie and others},
  journal={arXiv preprint arXiv:2508.18265},
  year={2025}
}

@misc{kimiteam2025kimivltechnicalreport,
      title={{Kimi-VL} Technical Report}, 
      author={Kimi Team and Angang Du and Bohong Yin and Bowei Xing and Bowen Qu and Bowen Wang and Cheng Chen and Chenlin Zhang and Chenzhuang Du and Chu Wei and Congcong Wang and Dehao Zhang and Dikang Du and Dongliang Wang and Enming Yuan and Enzhe Lu and Fang Li and Flood Sung and Guangda Wei and Guokun Lai and Han Zhu and Hao Ding and Hao Hu and Hao Yang and Hao Zhang and Haoning Wu and Haotian Yao and Haoyu Lu and Heng Wang and Hongcheng Gao and Huabin Zheng and Jiaming Li and Jianlin Su and Jianzhou Wang and Jiaqi Deng and Jiezhong Qiu and Jin Xie and Jinhong Wang and Jingyuan Liu and Junjie Yan and Kun Ouyang and Liang Chen and Lin Sui and Longhui Yu and Mengfan Dong and Mengnan Dong and Nuo Xu and Pengyu Cheng and Qizheng Gu and Runjie Zhou and Shaowei Liu and Sihan Cao and Tao Yu and Tianhui Song and Tongtong Bai and Wei Song and Weiran He and Weixiao Huang and Weixin Xu and Xiaokun Yuan and Xingcheng Yao and Xingzhe Wu and Xinxing Zu and Xinyu Zhou and Xinyuan Wang and Y. Charles and Yan Zhong and Yang Li and Yangyang Hu and Yanru Chen and Yejie Wang and Yibo Liu and Yibo Miao and Yidao Qin and Yimin Chen and Yiping Bao and Yiqin Wang and Yongsheng Kang and Yuanxin Liu and Yulun Du and Yuxin Wu and Yuzhi Wang and Yuzi Yan and Zaida Zhou and Zhaowei Li and Zhejun Jiang and Zheng Zhang and Zhilin Yang and Zhiqi Huang and Zihao Huang and Zijia Zhao and Ziwei Chen},
      year={2025},
      eprint={2504.07491},
      archivePrefix={arXiv},
      primaryClass={cs.CV},
      url={https://arxiv.org/abs/2504.07491}, 
}

@article{sellergren2025medgemma,
  title={MedGemma Technical Report},
  author={Sellergren, Andrew and Kazemzadeh, Sahar and Jaroensri, Tiam and Kiraly, Atilla and Traverse, Madeleine and Kohlberger, Timo and Xu, Shawn and Jamil, Fayaz and Hughes, Cían and Lau, Charles and others},
  journal={arXiv preprint arXiv:2507.05201},
  year={2025}
}

@article{pichai2025new,
  title={A new era of intelligence with gemini 3},
  author={Pichai, Sundar and Hassabis, Demis and Kavukcuoglu, Koray},
  journal={Mountain View, CA: Google},
  year={2025}
}

@article{singh2025openai,
  title={Openai gpt-5 system card},
  author={Singh, Aaditya and Fry, Adam and Perelman, Adam and Tart, Adam and Ganesh, Adi and El-Kishky, Ahmed and McLaughlin, Aidan and Low, Aiden and Ostrow, AJ and Ananthram, Akhila and others},
  journal={arXiv preprint arXiv:2601.03267},
  year={2025}
}

@misc{anthropic2026claude47,
  title  = {Introducing {Claude} 4.7},
  author = {{Anthropic}},
  year   = {2026},
  url    = {https://www.anthropic.com/news/claude-4-7},
  note   = {Accessed: \today}
}

@article{hoffman2014no,
  title={The No-U-Turn sampler: adaptively setting path lengths in Hamiltonian Monte Carlo.},
  author={Hoffman, Matthew D and Gelman, Andrew and others},
  journal={J. Mach. Learn. Res.},
  volume={15},
  number={1},
  pages={1593--1623},
  year={2014}
}

@article{phan2019composable,
  title={Composable effects for flexible and accelerated probabilistic programming in NumPyro},
  author={Phan, Du and Pradhan, Neeraj and Jankowiak, Martin},
  journal={arXiv preprint arXiv:1912.11554},
  year={2019}
}

@article{bradbury2018jax,
  title={JAX: composable transformations of Python+ NumPy programs},
  author={Bradbury, James and Frostig, Roy and Hawkins, Peter and Johnson, Matthew James and Leary, Chris and Maclaurin, Dougal and Necula, George and Paszke, Adam and VanderPlas, Jake and Wanderman-Milne, Skye and others},
  year={2018}
}

@article{birnbaum1968some,
  title={Some latent trait models and their use in inferring an examinee's ability},
  author={Birnbaum, Allan},
  journal={Statistical theories of mental test scores},
  year={1968},
  publisher={Addison-Wesley}
}

@book{embretson2025item,
  title={Item response theory: Foundations for psychologists and social scientists},
  author={Embretson, Susan E and Reise, Steven P},
  year={2025},
  publisher={Routledge}
}

@inproceedings{mitchell2019model,
  title={Model cards for model reporting},
  author={Mitchell, Margaret and Wu, Simone and Zaldivar, Andrew and Barnes, Parker and Vasserman, Lucy and Hutchinson, Ben and Spitzer, Elena and Raji, Inioluwa Deborah and Gebru, Timnit},
  booktitle={Proceedings of the conference on fairness, accountability, and transparency},
  pages={220--229},
  year={2019}
}

@inproceedings{raji2022fallacy,
  title={The fallacy of AI functionality},
  author={Raji, Inioluwa Deborah and Kumar, I Elizabeth and Horowitz, Aaron and Selbst, Andrew},
  booktitle={Proceedings of the 2022 ACM conference on fairness, accountability, and transparency},
  pages={959--972},
  year={2022}
}

@article{egressy2026set,
  title={Set-llm: A permutation-invariant llm},
  author={Egressy, Beni and St{\"u}hmer, Jan},
  journal={Advances in Neural Information Processing Systems},
  volume={38},
  pages={62798--62834},
  year={2026}
}

@article{zhao2024wildchat,
  title={Wildchat: 1m chatgpt interaction logs in the wild},
  author={Zhao, Wenting and Ren, Xiang and Hessel, Jack and Cardie, Claire and Choi, Yejin and Deng, Yuntian},
  journal={arXiv preprint arXiv:2405.01470},
  year={2024}
}

@inproceedings{chou2025visionarena,
  title={Visionarena: 230k real world user-vlm conversations with preference labels},
  author={Chou, Christopher and Dunlap, Lisa and Mashita, Koki and Mandal, Krishna and Darrell, Trevor and Stoica, Ion and Gonzalez, Joseph E and Chiang, Wei-Lin},
  booktitle={Proceedings of the IEEE/CVF Conference on Computer Vision and Pattern Recognition},
  pages={3877--3887},
  year={2025}
}
